\pdfoutput=1
\documentclass[11pt]{article}
\usepackage{xcolor}
\usepackage{emnlp2022}
\usepackage{times}
\usepackage{latexsym}
\usepackage[T1]{fontenc}
\usepackage[utf8]{inputenc}
\usepackage{microtype}

\usepackage{makecell}
\usepackage{booktabs}
\usepackage{multirow}
\usepackage{multicol}
\usepackage{subfig}
\usepackage{appendix}

\usepackage{url}

\usepackage{xspace}
\usepackage{mathtools}
\usepackage{amssymb}
\usepackage{pifont}
\usepackage{footnote}
\usepackage{tablefootnote}
\usepackage{graphicx}
\usepackage{color, colortbl}
\usepackage{xstring}
\usepackage{comment}
\usepackage{float}
\restylefloat{table}
\usepackage{array,booktabs,makecell}

\usepackage[normalem]{ulem}

\usepackage{arydshln}

\usepackage{titlesec}

 \usepackage{color,soul}

\def \ourLID{AfroLID}

\newcommand\blfootnote[1]{%
  \begingroup
  \renewcommand\thefootnote{}\footnote{#1}%
  \addtocounter{footnote}{-1}%
  \endgroup
}
\title{\ourLID: A Neural Language Identification Tool for African Languages}

\author{\normalsize Ife Adebara$^{1,\star}$ ~ AbdelRahim Elmadany$^{1,\star}$ ~ Muhammad Abdul-Mageed$^{1,2}$ ~ Alcides Alcoba Inciarte$^{1}$ \\
\normalsize $^{1}$Deep Learning \& Natural Language Processing Group,
  The University of British Columbia\\\normalsize  $^{2}$Department of Natural Language Processing \& Department of Machine Learning, MBZUAI\\ %
  \texttt{\normalsize \{ife.adebara@,a.elmadany@,muhammad.mageed@,alcobaaj@mail.\}ubc.ca}}
  
\begin{document}
\maketitle
\begin{abstract}
Language identification (LID) is a crucial precursor for NLP, especially for mining web data. Problematically, most of the world's $7000$+ languages today are not covered by LID technologies. We address this pressing issue for Africa by introducing~\ourLID, a neural LID toolkit for $517$ African languages and varieties.~\ourLID~exploits a multi-domain web dataset manually curated from across $14$ language families utilizing five orthographic systems. When evaluated on our blind Test set,~\ourLID~achieves $95.89$ $F_1$-score. We also compare~\ourLID~to five existing LID tools that each cover a small number of African languages, finding it to outperform them on most languages. We further show the utility of~\ourLID~in the wild by testing it on the acutely under-served Twitter domain. Finally, we offer a number of controlled case studies and perform a linguistically-motivated error analysis that allow us to both showcase~\ourLID's powerful capabilities and limitations.\blfootnote{ $^{\star}$ Authors contributed equally.}\footnote{\ourLID~is publicly available at \href{https://github.com/UBC-NLP/afrolid}{https://github.com/UBC-NLP/afrolid}.
} 
\end{abstract}

\section{Introduction}\label{sec:intro}
\begin{figure}[t]
  \centering
  \includegraphics[width=\columnwidth]{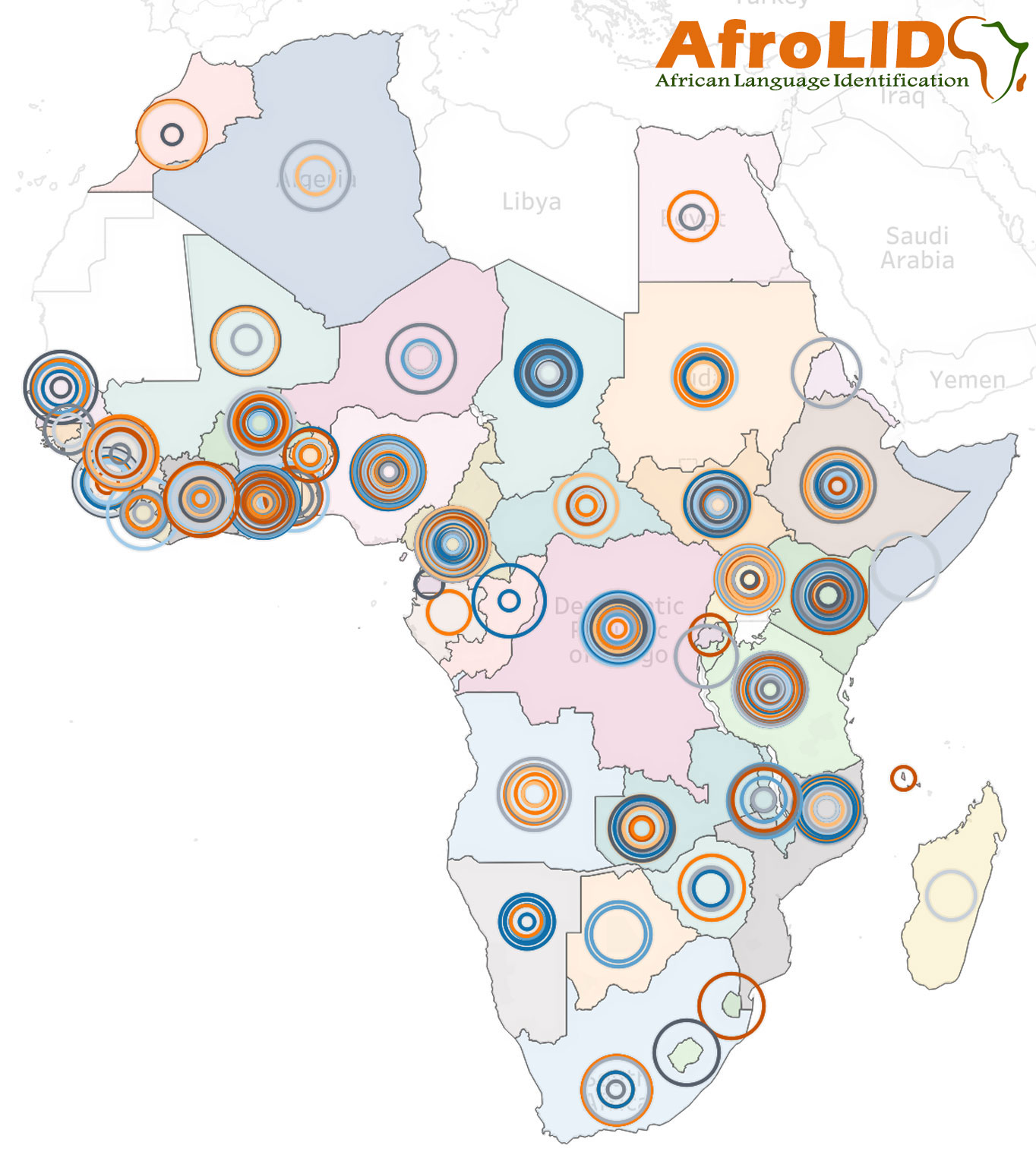}
\caption{\small  All $50$ African countries in our data, with our $517$ languages/language varieties in colored circles overlayed within respective countries. More details are in Appendix~\ref{sec:apx_covered_langs}.}
\label{fig:countries} 
\end{figure}




Language identification (LID) is the task of identifying the human language a piece of text or speech segment belongs to. The proliferation of social media have allowed greater access to multilingual data, making automatic LID an important first step in processing human language appropriately~\cite{tjandraetal2021improved, thara_etal_2021}. This includes applications in speech, sign language, handwritten text, and other modalities of language. It also includes distinguishing languages in code-mixed datasets~\cite{mageed2020microdialects, thara_etal_2021}. Unfortunately, for the majority of languages in the world, including most African languages, we do not have the resources for developing LID tools.

This situation has implications for the future NLP technologies. For instance, LID has facilitated development of widely multilingual models such mT5 \cite{xue-etal-2021-mt5} and large multilingual datasets such as CCAligned~\cite{elkishky_ccaligned_2020}, ParaCrawl~\cite{espla2019paracrawl}, WikiMatrix~\cite{schwenk2019wikimatrix}, OSCAR~\cite{suarez2020monolingual}, and mC4~\cite{xue-etal-2021-mt5} which have advanced research in NLP. Comparable resources are completely unavailable for the majority of the world's $7000$+ today, with only poor coverage of the so-called low-resource languages (LR). This is partly due to absence of LID tools, and impedes future NLP progress on these languages~\cite{adebara-abdul-mageed-2022-towards}. The state of African languages is not any better than other regions:~\newcite{caswell2021quality} perform a manual evaluation of $205$ datasets involving African languages such as those in CCAligned, ParaCrawl, WikiMatrix, OSCAR, and mC4 and show that at least $15$ corpora were completely erroneous, a significant fraction contained less than $50\%$ of correct data, and $82$ corpora were mislabelled or used ambiguous language codes. These consequently affect the quality of models built with these datasets.~\newcite{alabi2020massive} find that $135$K out of $150$K words in the fastText embeddings for Yor\`{u}b\'{a} belong to other languages such as English, French, and Arabic. New embedding models created by~\newcite{alabi2020massive} with a curated high quality dataset outperform off-the-shelf fastText embeddings, even though the curated data is smaller.


In addition to resource creation, lack (or poor performance) of LID tools negatively impacts preprocessing of LR languages since LID can be a prerequisite for determining, e.g.,  appropriate tokenization.~\cite{duvenhage2017improved}. Furthermore, some preprocessing approaches may be necessary for certain languages, but may hurt perforrmance in other languages~\cite{adebara-abdul-mageed-2022-towards}. Developing LID tools is thus vital for all NLP. In this work, we focus on LID for African languages and introduce~\ourLID.  

\begin{figure}[ht]
  \centering
  \includegraphics[width=\columnwidth]{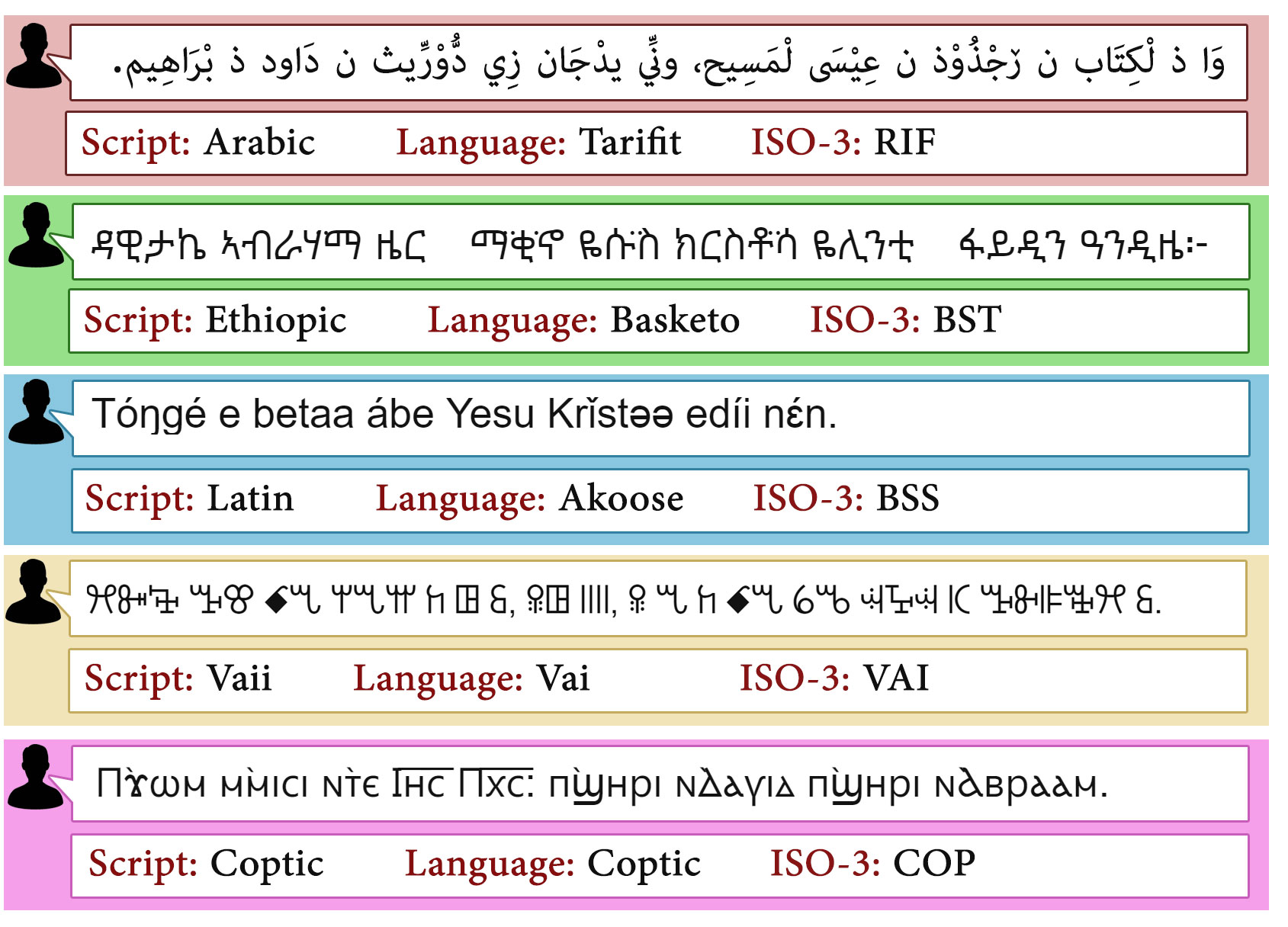}
\caption{\small Examples from the five scripts in our data.}
\label{fig:script_examples} 
\end{figure}

~\ourLID~is a neural LID tool that covers $517$ African languages and language varieties\footnote{Our dataset involves different forms that can arguably be viewed as varieties of the same language such as Twi and Akan.} across $14$ language families. The languages covered belong to $50$ African countries and are written in five diverse scripts. We show the countries covered by \ourLID~in Figure~\ref{fig:countries}. Examples of the different scripts involved in the $517$ languages are displayed in Figure~\ref{fig:script_examples}. To the best of our knowledge, \ourLID~supports the \textit{largest} subset of African languages to date. \ourLID~is also usable without any end-user training, and it exploits data from a variety of domains to ensure robustness. We manually curate our clean training data, which is of special significance in low resource settings. We show the utility of \ourLID~in the wild by applying it on two Twitter datasets and compare its performance with existing LID tools that cover any number of African languages such as CLD2~\cite{mccandless2010accuracy}, CLD3~\cite{salcianu2018compact}, \href{https://github.com/wooorm/franc}{Franc}, LangDetect~\cite{nakatani2010langdetect}, and Langid.py~\cite{lui-baldwin-2012-langid}. Our results show that \ourLID~consistently outperforms \textit{all} other LID tools for almost all languages, and serves as the new SOTA for language identification for African languages. 

To summarize, we offer the following main contributions:
\begin{enumerate}
    \item We develop~\ourLID, a SOTA LID tool for $517$ African languages and language varieties. To facilitate NLP research, we make our models publicly available.
    \item We carry out a study of LID tool performance on African languages where we compare our models in controlled settings with several tools such as CLD2, CLD3, Franc, LangDetect, and Langid.py.
    \item Our models exhibit highly accurate performance in the wild, as demonstrated by applying~\ourLID~on Twitter data.
    \item We provide a wide range of controlled case studies and carry out a linguistically-motivated error analysis of~\ourLID. This allows us to motivate plausible directions for future research, including potentially beyond African languages.
\end{enumerate}

The rest of the paper is organized as follows: In Section~\ref{sec:langs} we discuss a number of typological features of our supported languages. We describe~\ourLID's training data in Section~\ref{sec:africanData}. Next, we introduce~\ourLID~in~\ref{sec:afroLID}. This includes our experimental datasets and their splits, preprocessing, vocabulary, implementation and training details, and our evaluation settings. We present performance of~\ourLID~in Section~\ref{sec:perform_analysis} and compare it to other LID tools. Our analysis show that \ourLID~outperforms other models for most languages. In the same section, we also describe the utility of \ourLID~on non-Latin scripts, Creole languages, and languages in close geographical proximity. Although \ourLID~is not trained on Twitter data, we experiment with tweets in Section~\ref{sec:twitter} in order to investigate performance of \ourLID~in out of domain scenarios. Through two diagnostic studies, we demonstrate \ourLID's robustness. We provide an overview of related work in Section~\ref{sec:litreview}. We conclude in Section~\ref{sec:conclusion}, and outline a number of limitations for our work in Section~\ref{sec:limitations}.


\section{Typological Information}\label{sec:langs}
\noindent{\textbf{Language Families}.} We experiment with $517$ African languages and language varieties across $50$ African countries. These languages belong to $14$ language families \cite{ethnologue} as follows: Afro-Asiatic, Austronesian, Creole (English based), Creole (French based), Creole (Kongo based), Creole (Ngbadi based), Creole (Portuguese based), Indo-European, Khoe-Kwadi (Hainum), Khoe-Kwadi (Nama), Khoe-Kwadi (Southwest), Niger-Congo, and Nilo-Saharan. The large and typologically diverse data we exploit hence endow our work with wide coverage. We show in Figure \ref{fig:countries} a map of Africa with the countries~\ourLID~covers. We also show the number of languages we cover, per country, in Figure~\ref{sec:apx_covered_langs} in the Appendix. Table~\ref{tab:lang_listI}, Table~\ref{tab:lang_listII}, and Table~\ref{tab:lang_listIII} in the Appendix also provide a list of the languages~\ourLID~handles. We represent the languages using ISO-3 codes\footnote{\href{https://glottolog.org/glottolog/language}{https://glottolog.org/glottolog/language}.} for both individual languages and macro-languages. We use a macro-language tag when the language is known but the specific dialect is unknown. For this reason we specify that~\ourLID~supports $517$ African languages and language varieties. 







\noindent{\textbf{Sentential Word Order}.}
There are seven categories of word order across human languages around the world. These are subject-verb-object (SVO), subject-object-verb (SOV), object-verb-subject (OVS), object-subject-verb (OSV), verb-object-subject (VOS), verb-subject-object (VSO), and languages lacking a dominant order (which often have a combination of two or more orders within its grammar)~\cite{wals}. Again, our dataset is very diverse: we cover five out of these seven types of word order. Table~\ref{tab:orders} shows sentential word order in our data, with some representative languages for each category.

\begin{table}[!ht]
\begin{center}
\small 
\centering
\setlength{\tabcolsep}{5pt}
\begin{tabular}{ll}
 \toprule
\textbf{Word Order} & \textbf{Example Languages} \\ 
 \midrule
 SVO & Xhosa, Zulu, Y{o}r\`{u}b\'{a} \\
 SOV & Khoekhoe, Somali, Amharic  \\
 VSO & Murle, Kalenjin   \\
 VOS & Malagasy  \\
 No-dominant-order & Siswati, Nyamwezi, Bassa \\
 \bottomrule
\end{tabular}
\end{center}
\caption{Sentential word order in our data.}\label{tab:orders}
\end{table}

\noindent{\textbf{Diacritics}.}
Diacritic marks are used to overcome the inadequacies of an alphabet in capturing important linguistic information by adding a distinguishing mark to a character in an alphabet. Diacritics are often used to indicate tone, length, case, nasalization, or even to distinguish different letters of a language's alphabet~\cite{Wells2000OrthographicDA, hyman2003african,creissels2008africa}. Diacritics can be placed above, below or through a character. Diacritics are common features of the orthographies of African languages. Out of $517$ languages/language varieties in our training data, $295$ use some diacritics in their orthographies. 
We also provide a list of languages with diacritics in our training data in Table~\ref{tab:diacritics_in_training} in the Appendix.  

\begin{table}[!ht]
\begin{center}
\small 
\centering
\setlength{\tabcolsep}{5pt}
\begin{tabular}{ll}
 \toprule
\multicolumn{1}{l}{\textbf{Script}} & \multicolumn{1}{l}{\textbf{Languages }} \\ 
 \midrule
 Ethiopic & Amharic, Basketo, Maale, \\
 & \textsuperscript{$\star$}Oromo, Sebat Bet Gurage \\
& Tigrinya, Xamtanga\\\hdashline
 Arabic & Fulfude Adamawa, Fulfude Caka \\ 
 & Tarifit \\\hdashline
 Vai & Vai   \\\hdashline
 Coptic & Coptic \\
 \bottomrule
\end{tabular}
\end{center}
\caption{Non-Latin scripts in AfroLID data. \textbf{\textsuperscript{$\star$}Oromo}: is available in Latin script as well.}\label{tab:order}
\end{table}
\noindent{\textbf{Scripts}.}
Our dataset consists of $14$ languages written in four different non-Latin scripts and $499$ languages written in Latin scripts. The non-Latin scripts are Ethiopic, Arabic, Vai, and Coptic.



\section{Curating an African Language Dataset}\label{sec:africanData}
\ourLID~is trained using a multi-domain, multi-script language identification dataset that we manually curated for building our tool. To collect the dataset, we perform an extensive manual analysis of African language presence on the web, identifying as much publicly available data from the $517$ language varieties we treat as is possible. We adopt this manual curation approach since there are only few African languages that have any LID tool coverage. In addition, available LID tools that treat African languages tend to perform unreliably~\cite{caswell2021quality}. We therefore consult research papers focusing on African languages, such as ~\cite{adebara-abdul-mageed-2022-towards}, or provide language data~\cite{muhammad2022naijasenti, alabi2020massive}, sifting through references to find additional African data sources. Moreover, we search for newspapers across all $54$ African countries.\footnote{\href{https://www.worldometers.info/geography/how-many-countries-in-africa/}{https://www.worldometers.info/geography/how-many-countries-in-africa/}.} We also collect data from social media such as blogs and web fora written in African languages as well as databases that store African language data. These include~\href{https://lanfrica.com/records?task=natural\%20language\%20processing,machine\%20translation\&page=2}{LANAFRICA},~\href{https://repo.sadilar.org/discover}{SADiLaR},~\href{https://github.com/masakhane-io}{Masakhane},~\href{https://github.com/Niger-Volta-LTI}{Niger-Volta-LTI}, and~\href{http://www.alt-i.org/about/}{ALTI}. Our resulting multi-domain dataset contains religious texts, government documents, health documents, crawls from curated web pages, news articles, and existing human-identified datasets for African languages. As an additional sanity check, we ask a number of native speakers from a subset of the languages to verify the correctness of the self-labels assigned in respective sources within our collections.\footnote{We had access to native speakers of Afrikaans, Y{o}r\`{u}b\'{a}, Igbo, Hausa, Luganda, Kinyarwanda, Chichewa, Shona, Somali, Swahili, Xhosa, Bemba, and Zulu.} Our manual inspection step gave us confidence about the quality of our dataset, providing near perfect agreement by native speakers with labels from data sources. In total, we collect $~100$ million sentences in $528$ languages across $14$ language families in Africa and select $517$ languages which had at least $2000$ sentences. Again, the dataset has various orthographic scripts, including $499$ languages in Latin scripts, eight languages in Ethiopic scripts, four languages in Arabic scripts, one language in Vai scripts, and one in Coptic scripts. 
\section{\ourLID}\label{sec:afroLID}

\noindent{\textbf{Experimental Dataset and Splits}.} 
From our manually-curated dataset, we randomly select $5,000$, $50$, and $100$ sentences for train, development, and test, respectively, for each language.\footnote{We remove languages with data less than $2,000$ sentences, as explained earlier.} Overall, \ourLID~data comprises $2,496,980$ sentences for training (Train), $25,850$ for development (Dev), and $51,400$ for test (Test) for $517$ languages and language varieties. 

\noindent{\textbf{Preprocessing}.} We ensure that our data represent naturally occurring text by performing only minimal preprocessing. Specifically, we tokenize our data into character, byte-pairs, and words. We do not remove diacritics and use both precomposed and decomposed characters to cater for the inconsistent use of precomposed and decomposed characters by many African languages in digital media.\footnote{A Unicode entity that combines two or more other characters may be precomposed or decomposed. For example, \"{a} can be precomposed into $U+0061 U+0308$ or decomposed into $U+00E4$. In Unicode, they are included primarily to aid computer systems with incomplete Unicode support, where equivalent decomposed characters may render incorrectly.} We create our character level tokenization scripts and generate our vocabulary using \href{https://github.com/facebookresearch/fairseq}{Fairseq}. We use \href{https://github.com/google/sentencepiece}{sentencepiece tokenizer} for the word level and byte-pair tokens before we preprocess in Fairseq. 

\noindent{\textbf{Vocabulary}.} We experiment with byte-pair (BPE), word, and character level encodings. We used vocabulary sizes of $64$K, $100$K, and $2,260$ for the bpe, word, and character level models across the $517$ language varieties. The characters included both letters, diacritics, and symbols from other non-Latin scripts for the respective languages. 

\begin{figure}[ht]
  \centering
  \includegraphics[width=\linewidth]{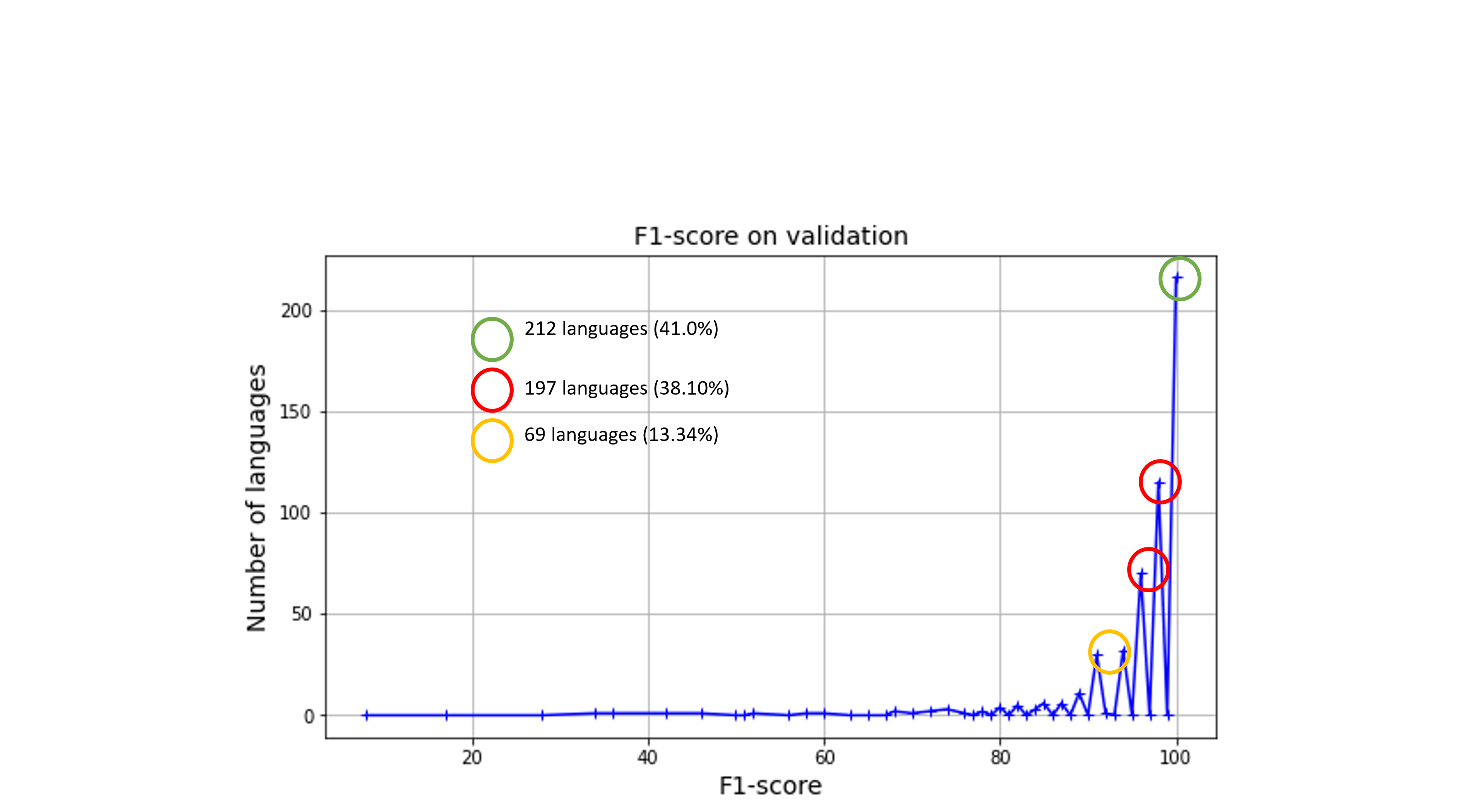}
\caption{\small $F_1$ distribution on \ourLID~Dev set.}\label{fig:line_chart_f1-score} 
\end{figure}

\noindent{\textbf{Implementation}.} \ourLID~is built using a Transformer architecture trained from scratch. We use $12$ attention layers with $12$ heads in each layer, $768$ hidden dimensions, making up $\sim200$M parameters.\footnote{This architecture is similar to XMLRBase~\cite{conneau-etal-2020-unsupervised}.} 

\noindent{\textbf{Hyperparameter Search and Training}.}
To identify our best hyperparameters, we use a subset of our training data and the full development set for our hyperparameter search. Namely, we randomly sample $200$ examples from each language in our training data to create a smaller train set,\footnote{This helps us limit GPU hours needed for hyperparameter search.} while using our full Dev set. We train for up to $100$ epochs, with early stopping. We search for the following hyperparameter values, picking bolded ones as our best: dropout rates from the set \textit{\{\textbf{0.1}, 0.2, 0.3, 0.4, 0.5\}}, learning rates from \textit{\{5e-5, \textbf{5e-6}\}}, and patience from \textit{\{\textbf{10}, 20, 30\}}.  Other hyperparameters are similar to those for XML-R \cite{conneau-etal-2020-unsupervised}. We perform hyperparameter search only with our character level model and use identified values with both the BPE and word models. 

\begin{figure}[h!]
  \centering
  \includegraphics[width=\linewidth]{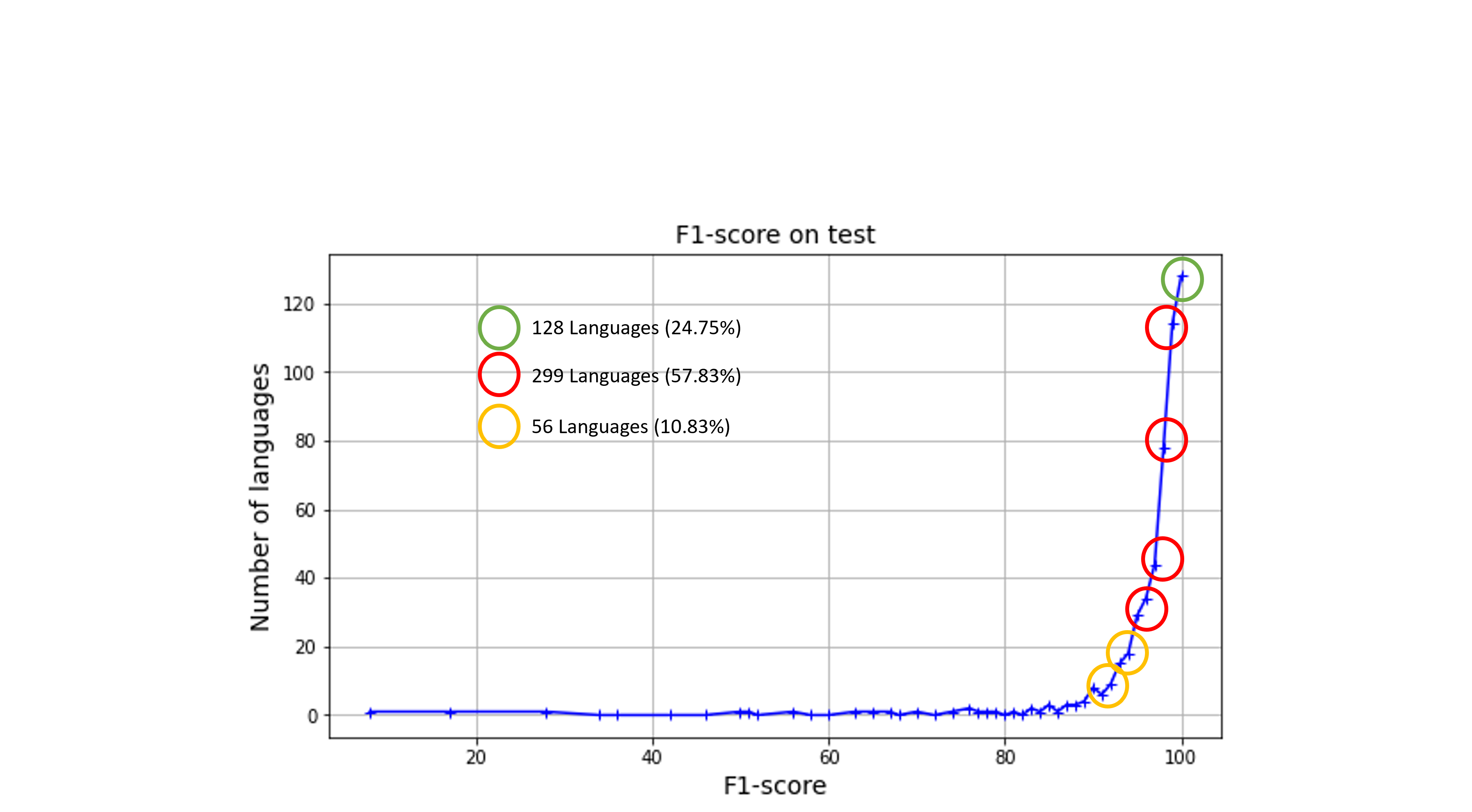}
\caption{\small $F_1$ distribution on \ourLID~Test set.}\label{fig:test_line_chart_f1-score} 
\end{figure}





\noindent{\textbf{Evaluation}.} We report our results in both macro $F_1$-score and accuracy, selecting our best model on Dev based on $F_1$. \textit{For all our models, we report the average of three runs}.

\section{Model Performance and Analysis}~\label{sec:perform_analysis}
As Table~\ref{tab:results_3_models} shows, our \textbf{BPE model} outperforms both the \textbf{char} and \textbf{word} models on both Dev and Test data. On Dev, our BPE model acquires \textbf{$96.14$} $F_1$ and \textbf{$96.19$} acc, compared to \textbf{$85.75$} $F_1$ and \textbf{$85.85$} for char model, and \textbf{$90.22$} $F_1$ and \textbf{$90.34$} acc for word model, respectively. Our BPE model similarly excels on Test, with \textbf{$95.95$} $F_1$ and \textbf{$96.01$} acc. We inspect the distribution of $F_1$ on the entire Dev and Test sets using our BPE model, as shown in Figures~\ref{fig:line_chart_f1-score} and ~\ref{fig:test_line_chart_f1-score}. As annotated on Figure~\ref{fig:line_chart_f1-score}, a total of $212$ languages out of the $517$ ($\%=41$) are identified with $100$ $F_1$, $197$ languages ($\%=38.10$) identified with $95$ and $99$ $F_1$, and $69$ languages ($\%=13.30$) identified with $90$--$95$ $F_1$. For Test data (Figure~\ref{fig:test_line_chart_f1-score}), on the other hand, $128$ ($\%=24.75$) languages are identified with $100$ $F_1$, $299$ languages ($\%=57.83$) are between $95$--$99$ $F_1$, while $56$ languages ($\%=10.83$) are between $90$--$95$ $F_1$.


\begin{table}[h!]
\centering
\small 
\centering
\setlength{\tabcolsep}{2pt}
\begin{tabular}{ccccc}
\toprule
\textbf{Model}                 & \textbf{Split} & $\bf F_1$\textbf{-score} & \textbf{Accuracy}  & \textbf{Checkpoint}  \\
\midrule
\multirow{2}{*}{Char} & Dev   & $85.75$ & $85.85$ & \multirow{2}{*}{$69$}   \\
                      & Test  & $81.20$  & $81.30$   &  \\
\hdashline
\multirow{2}{*}{BPE} & Dev   &\underline{$96.14$} & \underline{$96.19$} & \multirow{2}{*}{$73$}    \\ 
                      & Test  & $95.95$    & $96.01$ &   \\
\hdashline
\multirow{2}{*}{Word} & Dev   &  $90.22$ & $90.34$ & \multirow{2}{*}{$65$}      \\
                      & Test  & $89.04$  & $89.01$   & 
                      \\
\bottomrule
\end{tabular}
\caption{Results on the BPE, word level, and character level models. \textbf{Bolded}: best result on Test. \underline{Underlined}: best result on Dev.}
\label{tab:results_3_models}
\end{table}






\textbf{\ourLID~in Comparison}
Using our Dev and Test data, we compare our best \ourLID~model (BPE model) with the following LID tools: CLD2, CLD3, Franc, LangDetect, and Langid.py. Since these tools do not support all our \ourLID~languages, we compare accuracy and $F_1$-scores of our models only on languages supported by each of these tools. 
As Tables~\ref{tab:res} and \ref{tab:testres} show, \ourLID~outperforms other tools on $7$ and $8$ languages out of $16$ languages on the Dev set and Test set, respectively. We also compare $F_1$-scores of \textbf{Franc} on the $88$ African languages Franc supports with the $F_1$-scores of \ourLID~on those languages. As shown in Tables~\ref{tab:franc_vs_afrolid} and~\ref{tab:franc_vs_afrolid_test}, \ourLID~outperforms Franc on $78$ languages and has similar $F_1$-score on five languages on the Dev set. \ourLID~also outperforms Franc on $76$ languages, and has similar $F_1$-score on five languages on the Test set.
\begin{table}[h!]
\centering

\small 
\centering
\resizebox{\columnwidth}{!}{%
\begin{tabular}{lrrrrrr}

\toprule
\textbf{Lang.}  & \textbf{CLD2} & \textbf{CLD3} & \textbf{Langid.py} & \textbf{LangDetect} & \textbf{Franc} & \textbf{\ourLID}\\ 

\midrule
        afr & $94.00$ & $91.00$ & $69.00$ & $88.23$ & $81.00$ & \textbf{$\bf97.00$} \\ 
        amh & - & $97.00$ & \textbf{$\bf100.00$} & - & $35.00$ & $97.00$ \\ 
        hau & - & $83.00$ & - & - & $77.00$ & \textbf{$\bf88.00$} \\ 
        ibo & - & $96.00$ & - & - & $88.00$ & \textbf{$\bf97.00$} \\ 
        kin &\textbf{$\bf92.00$} & - & $45.00$ & - & $47.00$ & $89.00$ \\ 
        lug & \textbf{$84.00$} & - & - & - & $64.00$ & \textbf{$\bf87.00$} \\ 
        mlg & - & \textbf{$\bf100.00$} & $98.00$ & - & - & \textbf{$\bf100.00$}\\ 
        nya & - & \textbf{$\bf96.00$} & - & - & $75.00$ & \textbf{$92.00$} \\ 
        sna & - & {$\bf100.00$} & - & - & $91.00$ & \textbf{$97.00$} \\ 
        som & - & $92.00$ & - & - & $89.00$ & $\bf95.00$ \\ 
        sot & - & \textbf{$\bf99.00$} & - & - & $93.00$ & $88.00$ \\ 
        swa & $99.00$ & $91.00$ & $90.00$ & \textbf{$\bf100.00$} & - & $92.00$ \\ 
        swc & $93.00$ & $94.00$ & $96.00$ & \textbf{$\bf97.02$} & - & $87.00$ \\ 
        swh & $89.00$ & \textbf{$\bf92.00$} & $88.23$ & $87.19$ & $70.00$ & $77.00$ \\ 
        xho & - & $59.00$ & \textbf{$\bf88.00$} & - & $30.00$ & $67.00$ \\ 
        yor & - & $25.00$ & - & - & $66.00$ & \textbf{$\bf98.00$} \\ 
        zul & - & \textbf{$\bf89.00$} & $20.00 $ & - & $40.00$ & $50.00$ \\ 
\bottomrule
\end{tabular}%
}
\caption{A comparison of results on \ourLID~with CLD2, CLD3, Langid.py, LangDetect, and Franc using $F_1$-score on the Test set. $-$ indicates that the tool does not support the language.}
\label{tab:testres}
\end{table}

\begin{table*}[h!]
\resizebox{2\columnwidth}{!}{%
\begin{tabular}{lrr  lrr  lrr   lrr  lrr }
\toprule
{\textbf{ISO-3}} & {\textbf{AfroLID}} & {\textbf{Franc}} & {\textbf{ISO-3}} & {\textbf{AfroLID}} & {\textbf{Franc}} & {\textbf{ISO-3}} & {\textbf{AfroLID}} & {\textbf{Franc}} & {\textbf{ISO-3}} & {\textbf{AfroLID}} & {\textbf{Franc}} & {\textbf{ISO-3}} & {\textbf{AfroLID}} & {\textbf{Franc}} \\
\midrule
{ { aar}}      & { \textbf{100.00}}   & { 74.50}          & { { fat}}      & { \textbf{94.11}}   & { 88.23}          & { { koo}}      & { \textbf{96.07}}   & { 86.27}          & { { nso}}      & { \textbf{84.31}}   & { 70.58}          & { { tir}}      & { 98.03}   & {\textbf{100.00}}          \\
{ { ada}}      & { \textbf{98.03}}   & { 96.07}          & { { fon}}      & { \textbf{98.03}}   & { 86.27}          & { { kqn}}      & { \textbf{96.07}}   & { 86.27}          & { { nya}}      & { \textbf{96.07}}   & { 82.35}          & { { tiv}}      & { \textbf{100.00}}  & { 98.03}           \\
{ { afr}}      & { \textbf{94.11}}   & { 84.31}          & { { fuf}}      & { \textbf{98.03}}   & { 60.78}          & { { kqs}}      & { \textbf{100.00}}  & { 64.70}          & { { nym}}      & { \textbf{100.00}}  & { 52.94}          & { { toi}}      & { \textbf{100.00}}  & { 68.62}           \\
{ { amh}}      & { \textbf{98.03}}   & { 25.49}          & { { fuv}}      & { \textbf{90.19}}   & { 35.29}          & { { ktu}}      & { \textbf{96.07}}   & { 17.64}          & { { nyn}}      & { \textbf{92.15}}   & { 84.31}          & { { tsn}}      & { \textbf{70.58}}   & { 54.90}           \\
{ { bam}}      & { \textbf{70.58}}   & { 45.09}          & { { gaa}}      & { \textbf{96.07}}   & { \textbf{96.07}} & { { lia}}      & { \textbf{98.03}}   & { \textbf{98.03}} & { { nzi}}      & { \textbf{98.03}}   & { \textbf{98.03}} & { { tso}}      & { \textbf{96.07}}   & { 80.39}           \\
{ { bba}}      & { \textbf{98.03}}   & { 88.23}          & { { gaz}}      & { \textbf{96.07}}   & { 90.19}          & { { lin}}      & { \textbf{98.03}}   & { 96.07}          & { { pcm}}      & { \textbf{98.03}}   & { 78.43}          & { { twi}}      & { \textbf{90.19}}   & { 84.31}           \\
{ { bci}}      & { 76.47}            & { \textbf{86.27}} & { { gjn}}      & { \textbf{100.00}}  & { 94.11}          & { { lot}}      & { \textbf{100.00}}  & { 94.11}          & { { pov}}      & { \textbf{96.07}}   & { 86.27}          & { { umb}}      & { \textbf{90.19}}   & { 70.58}           \\
{ { bem}}      & { \textbf{82.35}}   & { 64.70}          & { { gkp}}      & { 64.70}            & { \textbf{68.62}} & { { loz}}      & { \textbf{96.07}}   & { 94.11}          & { { run}}      & { \textbf{84.31}}   & { 58.82}          & { { vai}}      & { \textbf{100.00}}  & { \textbf{100.00}} \\
{ { bfa}}      & { \textbf{100.00}}  & { 90.19}          & { { hau}}      & { \textbf{94.11}}   & { 82.35}          & { { lua}}      & { \textbf{98.03}}   & { 96.07}          & { { sag}}      & { \textbf{94.11}}   & { 17.64}          & { { ven}}      & { \textbf{96.07}}   & { \textbf{96.07}}  \\
{ { bin}}      & { 94.11}   & {\textbf{ 98.03} }         & { { ibb}}      & { \textbf{98.03}}   & { 86.27}          & { { lue}}      & { \textbf{90.19}}   & { 60.78}          & { { shk}}      & { \textbf{100.00}}  & { 96.07}          & { { vmw}}      & { \textbf{88.23}}   & { 80.39}           \\
{ { bum}}      & { \textbf{100.00}}  & { 52.94}          & { { ibo}}      & { \textbf{94.11}}   & { 90.19}          & { { lug}}      & { \textbf{86.27}}   & { 52.94}          & { { sna}}      & { \textbf{96.07}}   & { 80.39}          & { { wol}}      & { \textbf{68.62}}   & { 23.52}           \\
{ { cjk}}      & { \textbf{98.03}}   & { 52.94}          & { { kbp}}      & { \textbf{98.03}}   & { 94.11}          & { { lun}}      & { \textbf{98.03}}   & { 90.19}          & { { som}}      & { \textbf{98.03}}   & { 96.07}          & { { xho}}      & { \textbf{82.35}}   & { 64.70}           \\
{ { crs}}      & { \textbf{94.11}}   & { 82.35}          & { { kde}}      & { \textbf{96.07}}   & { 78.43}          & { { men}}      & { \textbf{98.03}}   & { 92.15}          & { { sot}}      & { \textbf{76.47}}   & { 90.19}          & { { xsm}}      & { \textbf{100.00}}  & { 25.49}           \\
{ { dag}}      & { \textbf{96.07}}   & { 96.07}          & { { kdh}}      & { \textbf{100.00}}  & { 92.15}          & { { mfq}}      & { \textbf{96.07}}   & { 01.96}           & { { ssw}}      & { \textbf{90.19}}   & { 84.31}          & { { yor}}      & { \textbf{100.00}}  & { 39.21}           \\
{ { dga}}      & { \textbf{100.00}}  & { 88.23}          & { { kea}}      & { \textbf{98.03}}   & { 3.92}           & { { mos}}      & { \textbf{94.11}}   & { 84.31}          & { { suk}}      & { \textbf{100.00}}  & { 31.37}          & { { zdj}}      & { \textbf{100.00}}  & { 62.74}           \\
{ { dip}}      & { \textbf{98.03}}   & { 84.31}          & { { kin}}      & { \textbf{80.39}}   & { 52.94}          & { { nba}}      & { \textbf{100.00}}  & { 56.86}          & { { sus}}      & { \textbf{100.00}}  & { 96.07}          & { { zul}}      & { \textbf{58.82}}   & { 37.25}           \\
{ { dyu}}      & { \textbf{98.03}}   & { 01.96}           & { { kmb}}      & { \textbf{100.00}}  & { 80.39}          & { { nbl}}      & { \textbf{80.39}}   & { 64.70}          & { { swh}}      & { \textbf{74.50}}   & { 72.54}          & { }               & { }                 & { }                \\
{ { ewe}}      & { 94.11}            & { \textbf{96.07}} & { { kng}}      & { \textbf{98.03}}   & { 66.66}          & { { ndo}}      & { \textbf{90.19}}   & { 82.35}          & { { tem}}      & { \textbf{96.07}}   & { 84.31}          & { }               & { }                 & { }
\\\hline
\multicolumn{4}{r}{\textbf{\ourLID~Average $F_1$-score}: $\bf93.21$ } &
\multicolumn{11}{r}{ \textbf{Franc Average $F_1$-score}: $72.85$ }
\\ \bottomrule
\end{tabular}%
}
\caption{$F_1$-scores on our Dev dataset for languages in \ourLID~and Franc for $88$ languages.}\label{tab:franc_vs_afrolid}
\end{table*}
\begin{table*}[h!]
\resizebox{2\columnwidth}{!}{%
\begin{tabular}{lrr lrr lrr  lrr lrr }
\toprule
{\textbf{ISO-3}} & {\textbf{AfroLID}} & {\textbf{Franc}} & {\textbf{ISO-3}} & {\textbf{AfroLID}} & {\textbf{Franc}} & {\textbf{ISO-3}} & {\textbf{AfroLID}} & {\textbf{Franc}} & {\textbf{ISO-3}} & {\textbf{AfroLID}} & {\textbf{Franc}} & {\textbf{ISO-3}} & {\textbf{AfroLID}} & {\textbf{Franc}} \\
\hline
{ { aar}}      & { \textbf{96.00}}  & { 74.00}         & { { fat}}      & { \textbf{98.00}}   & { 94.00}          & { {koo}}      & { \textbf{96.00}}   & {\textbf{96.00}}           & { { nso}}      & { \textbf{83.00}}   & { 59.00}          & { { tir}}      & { \textbf{99.00}}   & {97.00}          \\
{ { ada}}      & { \textbf{100.00}}   & { 98.00}          & { { fon}}      & { \textbf{97.00}}   & { 92.00}          & { { kqn}}      & { \textbf{98.00}}   & { 84.00}          & { { nya}}      & { \textbf{92.00}}   & { 75.00}          & { { tiv}}      & { \textbf{100.00}}  & {99.00}           \\
{ { afr}}      & { \textbf{97.00}}   & { 81.00}          & { { fuf}}      & { \textbf{93.00}}   & { 52.00}          & { { kqs}}      & { \textbf{95.00}}  & { 73.00}          & { { nym}}      & { \textbf{99.00}}  & { 54.00}          & { { toi}}      & { \textbf{98.00}}  & {80.00}           \\
{ { amh}}      & { \textbf{97.00}}   & { 36.00}          & { { fuv}}      & { \textbf{94.00}}   & { 61.00}          & { { ktu}}      & { \textbf{93.00}}   & { 19.00}          & { { nyn}}      & { \textbf{92.00}}   & { \textbf{92.00}}       & { { tsn}}      & { \textbf{76.00}}   & {33.00}           \\
{ { bam}}      & { \textbf{70.00}}   & { 30.00}          & { { gaa}}      & { 95.00}   & { \textbf{97.00}} & { { lia}}      & { 97.00}   & \textbf{{100.00}} & { { nzi}}      & { 97.00}   & { \textbf{98.00}} & { { tso}}      & { \textbf{99.00}}   & {94.00}           \\
{ { bba}}      & { \textbf{100.00}}   & { 83.00}          & { { gaz}}      & {94.00}   &{ \textbf{96.00}}          & { { lin}}      & { \textbf{99.00}}   & {98.00}         & { { pcm}}      & { \textbf{96.00}}   & {82.00}          & { { twi}}      & { \textbf{100.00}}   & {87.00}           \\
{ { bci}}      & \textbf{{98.00}}         & { 92.00} & { { gjn}}      & { 98.00}  & {\textbf{99.00}}           & { { lot}}      & { \textbf{99.00}}  & { 93.00}          & { { pov}}      & { \textbf{93.00}}   & {82.00}          & { { umb}}      & { \textbf{99.00}}   & {76.00}           \\
{ { bem}}      & { \textbf{94.00}}   & { 90.00}          & { { gkp}}      & { 63.00}            & { \textbf{69.00}} & { { loz}}      & { \textbf{95.00}}   & { 92.00}          & { { run}}      & { \textbf{91.00}}   & {68.00}          & { { vai}}      & { \textbf{100.00}}  & { \textbf{100.00}} \\
{ { bfa}}      & { \textbf{99.00}}  & { 91.00}          & { { hau}}      & { \textbf{88.00}}   & { 77.00}          & { { lua}}      & { \textbf{99.00}}   & { 87.00}          & { { sag}}      & { \textbf{100.00}}   & {30.00}          & { { ven}}      & { \textbf{95.00}}   & {85.00} \\
{ { bin}}      & { \textbf{99.00}}   & { 97.00}          & { { ibb}}      & { \textbf{98.00}}   & { 84.00}          & { { lue}}      & { \textbf{95.00}}   & { 68.00}          & { { shk}}      & { \textbf{100.00}}  & {93.00}          & { { vmw}}      & { \textbf{97.00}}   & { 95.00}           \\
{ { bum}}      & { \textbf{97.00}}  & { 72.00}          & { { ibo}}      & { \textbf{97.00}}   & { 88.00}          & { { lug}}      & { \textbf{87.00}}   & { 64.00}          & { { sna}}      & { \textbf{97.00}}   & {91.00}          & { { wol}}      & { \textbf{81.00}}   & {21.00}           \\
{ { cjk}}      & { \textbf{96.00}}   & { 56.00}          & { { kbp}}      & { \textbf{100.00}}   & {98.00}          & { { lun}}      & { \textbf{97.00}}   & { 86.00}          & { { som}}      & { \textbf{95.00}}   & {89.00}          & { { xho}}      & { \textbf{67.00}}   & {30.00}           \\
{ { crs}}      & { \textbf{96.00}}   & { 83.00}          & { { kde}}      & { \textbf{95.00}}   & { 60.00}          & { { men}}      & {98.00}   & { \textbf{99.00}}         & { { sot}}      & { 88.00}   & \textbf{ {93.00}  }        & { { xsm}}      & { \textbf{99.00}}  & {53.00}           \\
{ { dag}}      & { \textbf{100.00}}   & { \textbf{100.00}}          & { { kdh}}      & { \textbf{99.00}}  & { 95.00}          & { { mfq}}      & { \textbf{95.00}}   & {88.00}           & { { ssw}}      & { \textbf{86.00}}   & { 68.00}          & { { yor}}      & { \textbf{98.00}}  & {66.00}           \\
{ { dga}}      & { \textbf{100.00}}  & { 78.00}          & { { kea}}      & { \textbf{96.07}}   & { 0.00}           & { { mos}}      & { \textbf{97.00}}   & { 90.00}          & { { suk}}      & { \textbf{99.00}}  & {34.00}          & { { zdj}}      & { \textbf{96.00}}  & {63.00}           \\
{ { dip}}      & { \textbf{93.00}}   & {86.00}          & { { kin}}      & { \textbf{89.00}}   & { 47.00}          & { { nba}}      & { \textbf{99.00}}  & { 61.00}          & { { sus}}      & { \textbf{99.00}}  & {96.00}          & { { zul}}      & { \textbf{50.00}}   & {40.00}           \\
{ { dyu}}      & { \textbf{96.00}}   & { 00.00}           & { { kmb}}      & { \textbf{94.00}}  & { 71.00}          & { { nbl}}      & { \textbf{74.00}}   & { 47.00}          & { { swh}}      & { \textbf{77.00}}   & {70.00}          & { }               & { }                 & { }                \\
{ { ewe}}      & { \textbf{97.00} }           & { \textbf{97.00} } & { { kng}}      & { \textbf{98.00}}   & {58.00}          & { { ndo}}      & { \textbf{96.00}}   & { 76.00}          & { { tem}}      & { \textbf{99.00}}   & {88.00}          & { }               & { }                 & { }
\\ \hline
	
\multicolumn{4}{r}{\textbf{\ourLID~Average $F_1$-score}: $\bf91.63$ } &
\multicolumn{11}{r}{ \textbf{Franc Average $F_1$-score}: $74.81$ }
\\ \bottomrule
\end{tabular}%
}
\caption{$F_1$-scores on our Test dataset for languages in \ourLID~and Franc for $88$ languages.}
\label{tab:franc_vs_afrolid_test}
\end{table*}

\begin{figure}[h!]
  \centering
  \includegraphics[width=\linewidth]{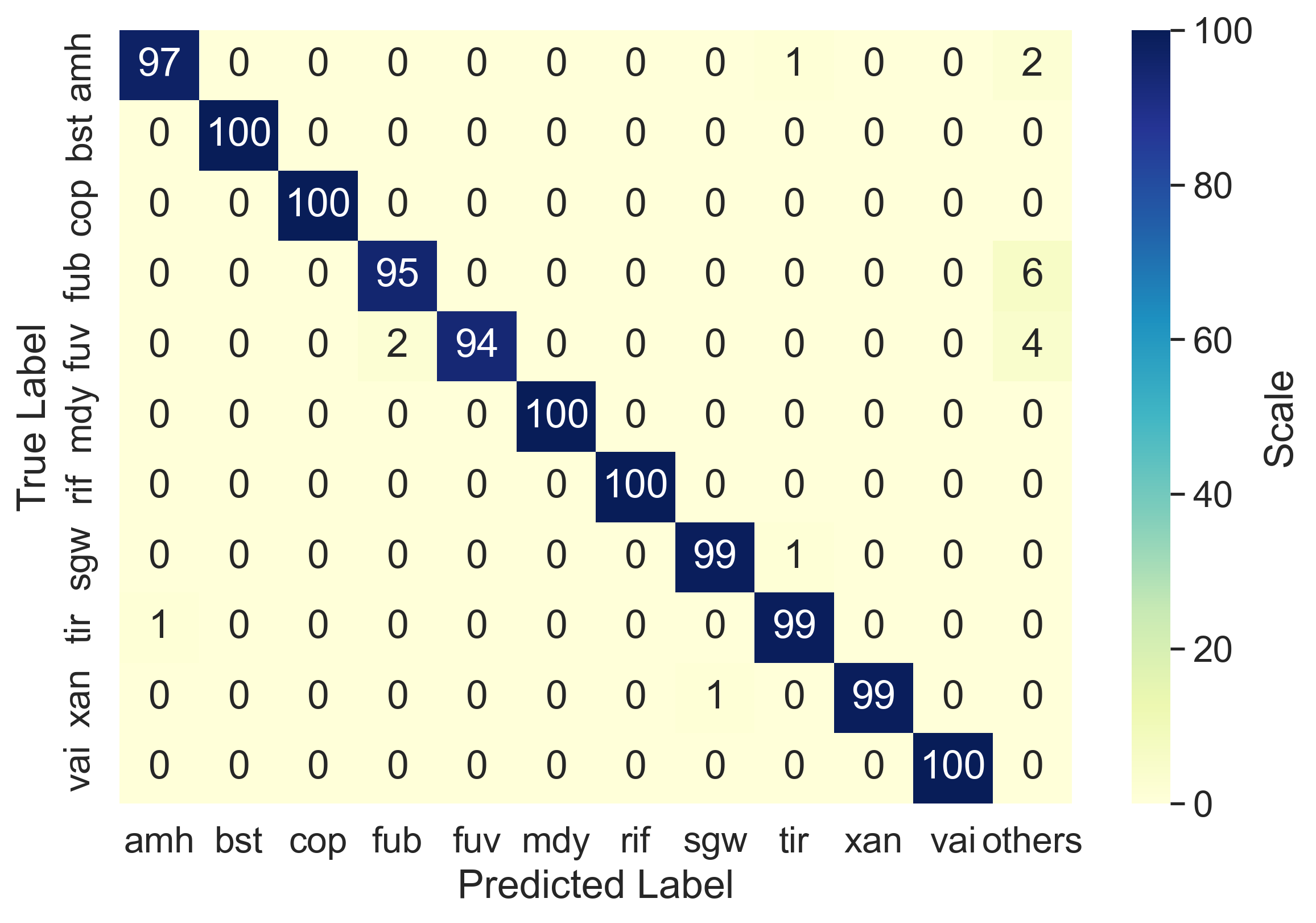}
\caption{\small Errors on the different script in \ourLID~Test set. We use ISO-3 codes to represent the languages. ``Others" refers to languages \ourLID~identifies as outside the list of languages selected for analysis.}\label{fig:s_confusion_matrix_test}
\end{figure}

\textbf{Effect of Non-Latin Script.}\label{subsubsec:nonlatin}
We investigate performance of \ourLID~on languages that use one of Arabic, Ethiopic, Vai, and Coptic scripts. Specifically, we investigate performance of \ourLID~on Amharic (amh), Basketo (bst), Maale (mdy), Sebat Bet Gurage (sgw), Tigrinya (tir), Xamtanga (xan), Fulfude Adamawa (fub), Fulfude Caka (fuv), Tarif (rif), Vai (vai), and Coptic (cop).\footnote{We do not investigate performance on Oromo because we had both Latin and Ethiopic scripts for Oromo in our training data.} Vai and Coptic, the two unique scripts in \ourLID~have an $F_1$-score of $100$ each. This corroborates research findings that languages written in unique scripts within an LID tool can be identified with up to $100\%$ recall, $F_1$-score, and/or accuracy even using a small training dataset~\cite{jauhiainen-etal-2017-evaluation}. We assume this to be the reason Langid.py outperforms \ourLID~on Amharic as seen in Table~\ref{tab:testres}, since Amharic is the only language that employs an Ethiopic script in langid.py. \ourLID, on the other hand, has $8$ languages using Ethiopic scripts. However, it is not clear why Basketo, which uses Ethiopic scripts has $100$ $F_1$-score. We, however, found errors in Amharic, Sebat Bet Gurage, and Xamtanga (which use Ethiopic scripts) as well as Fulfude Adamawa, and Fulfude Caka (which use Arabic scripts). We find that languages using Ethiopic scripts are often confused with those using Ethiopic scripts (except for $2\%$ of the time when Amharic is labelled as Wolof). We categorize this example under "others" in Figure \ref{fig:s_confusion_matrix_test} and \ref{fig:s_confusion_matrix}. On the other hand, Fulfude languages are wrongly labelled as other dialects of Fulfude that use Latin scripts. We visualize further details of the errors in Figure \ref{fig:s_confusion_matrix} (in Appendix) and \ref{fig:s_confusion_matrix_test} for our Dev and Test sets. 

\textbf{Creole Languages.}\label{subsubsec:creole}
We investigate performance of \ourLID~on Creole languages. Creole languages are vernacular languages that emerged as a result of trade interactions between speakers of mutually unintelligible languages~\cite{lent-etal-2022-ancestor}. A Creole language therefore shares lexical items and grammatical structures with one or more different, unrelated languages. As a result, Creole languages appear to be \textit{code-mixed}. \ourLID~is trained on nine Creole languages: Krio, Nigerian Pidgin, Cameroonian Pidgin, Seychelles Creole, Mauritian Creole, Kituba, Sango, Kabuverdianu, and Guinea-Bissau Creole. Krio, Cameroonian Pidgin, and Nigerian Pidgin are English based. Seychelles Creole and Mauritian Creole are French based. Kituba is Kongo based and Sango is Ngbadi based. Kabuverdianu and Guinea-Bissau Creole are Portuguese based. Evaluating \ourLID~on Creoles thus demonstrates the robustness of our model, since (as mentioned above) Creoles can be viewed as a type of \textit{code-mixed} language. We show performance of \ourLID~on the nine Creole languages in Figure \ref{fig:creole_conf_matrix} (in Appendix) and \ref{fig:creole_conf_matrix_test} for Dev and Test sets respectively. 

\begin{figure}[h!]
  \centering
  \includegraphics[width=\linewidth]{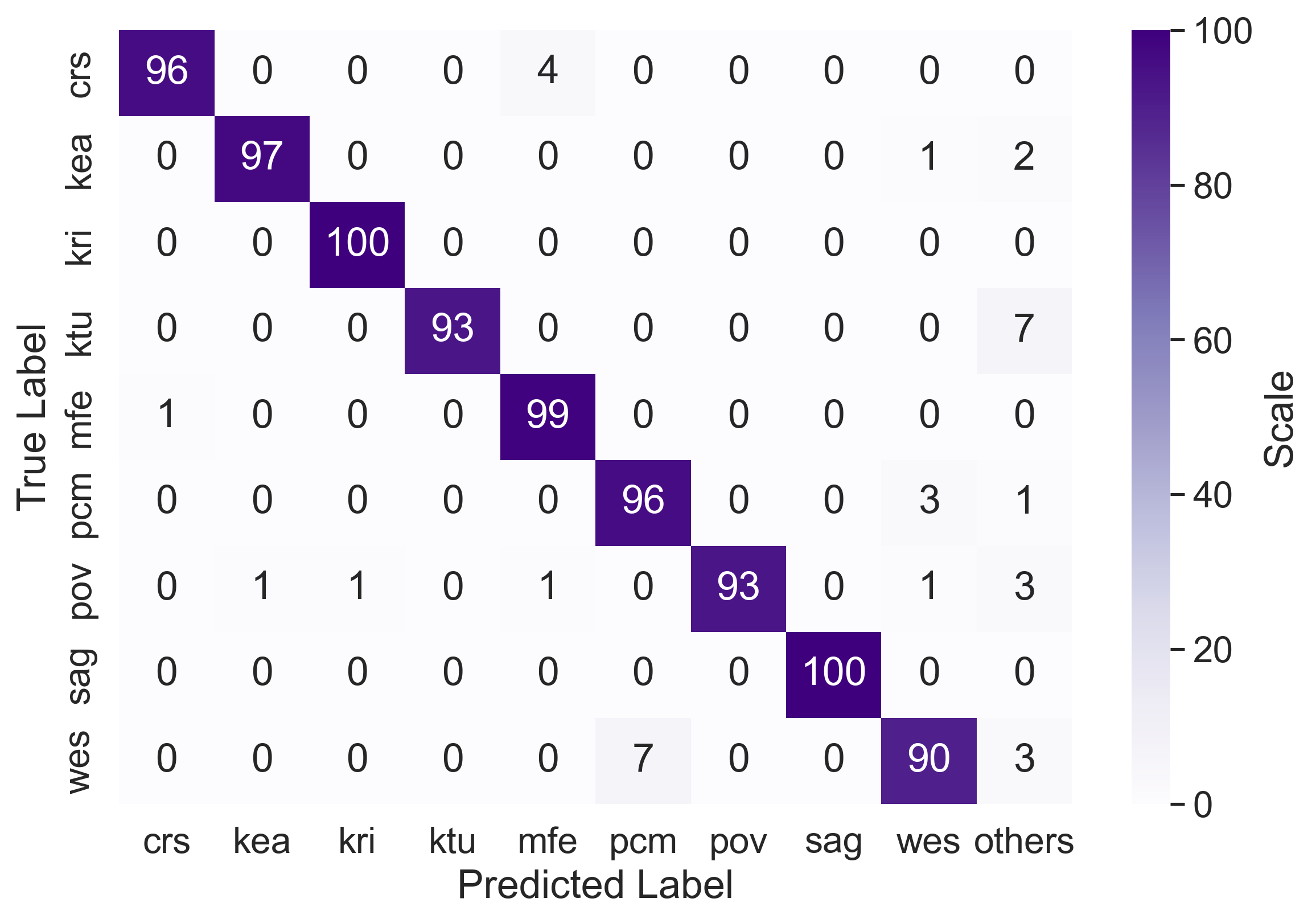}
\caption{\small Errors on the different Creoles in \ourLID. We use ISO-3 codes to represent the languages. ``Others" refers to languages \ourLID~identifies as outside the list of languages selected for analysis.}\label{fig:creole_conf_matrix_test} 
\end{figure}

We find that Guinea-Bissau Creole (pov), which is Portuguese based, is wrongly labelled as Kabuverdianu (kea) another Portuguese based Creole $1\%$ of the time. Cameroonian pidgin (wes) is also wrongly labelled as Nigerian pidgin (pcm) $7\%$ of the time. Since both Cameroonian and Nigerian Pidgin are English based, we assume lexical and/or grammatical similarities are responsible for these errors. It is also interesting to find cases where the wrong labels are languages spoken in the same geographical regions as the Creoles. For example, Kituba is wrongly labelled as Yombe, and both languages are spoken in Congo. Mauritian Creole (mfe), which is French based, is also wrongly labelled as Seychelles Creole (crs, another French based Creole) and two Indigenous languages spoken in Francophone Africa Ngiemboon, and Masana. We now further investigate the role of geographical proximity in our results.

\textbf{Effect of Geographic Proximity.}\label{subsubsec:geography}
We evaluate performance of \ourLID~on languages that share a large number of lexical items, or those that are spoken within the same country. In this analysis, we focus on $10$ South African languages: Afrikaans (afr), Ndebele (nbl), Sepedi (nso), Sotho (sot), Swati (ssw), Tswana (tsn), Tsonga (tso), Tsivenda (ven), Xhosa (xho), and Zulu (zul). We select South Africa because most South Africans are multi-lingual, and it is not uncommon to find code-mixing using a combination of Indigenous languages within the same text~\cite{FINLAYSONSLABBERT+1997+65+98, JESR5628}. Figures~\ref{fig:sa_conf_matrix} (in Appendix) and \ref{fig:sa_conf_matrixtest} show the types of errors \ourLID~makes in identifying these languages on our Dev and Test datasets respectively. We find that about $\sim70\%$ of the errors are with other South African languages. Another $16\%$ are with dialects from neighbouring countries including Tswa, a dialect of Tsonga, Ndebele (Zimbabwe) similar to Zulu, and Ronga, a dialect of Tsonga.\footnote{A total of $14\%$ of the errors are for other languages not related to South African languages.} We now provide a number of case studies we carry out to further probe \ourLID~performance.  

\begin{figure}[h!]
  \centering
  \includegraphics[width=\linewidth]{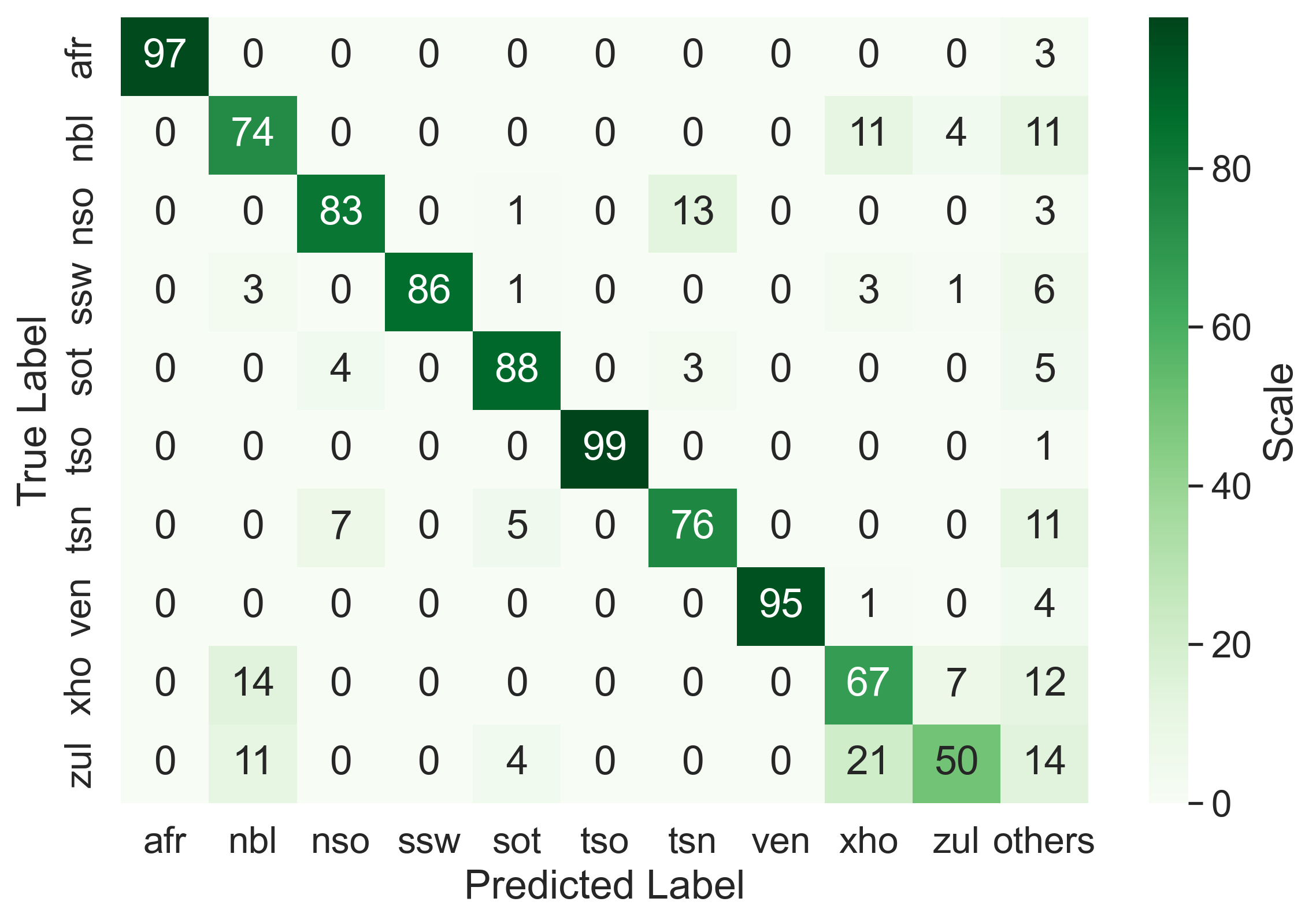}
\caption{\small Errors on Indigenous South African languages in \ourLID~Test data. ``Others" refers to languages \ourLID~identifies as outside the list of languages selected for analysis.}\label{fig:sa_conf_matrixtest}

\end{figure}




\begin{table*}[h!]
\setlength{\tabcolsep}{4pt}
\resizebox{2\columnwidth}{!}{%
\begin{tabular}{lllll}
\toprule
\textbf{Tool}  & \textbf{Covered/All} & \textbf{Training Data} & \textbf{Methodology} \\
\midrule
Langid.py& $7$/$97$   &  GDoc, SDoc, News, ENC, IC  &               Naive Bayes, $n$-gram         \\ 
Langdetect & $3/49$ & Wikipedia& Naive Bayes, char $n$-gram\\
CLD2 & $4/80$ & Unknown&  Naïve Bayes \\ 
CLD3 & $13/107$  & Unknown &  Neural network, char $n$-gram & \\
\href{https://github.com/davidjurgens/equilid}{Equilid} & $1/70$ & Several GDoc, SDoc, RDoc, News, ENC, IC, Twitter & Neural seq2seq \\
Fasttext & $5/176$ & \href{https://www.wikipedia.org/}{Wiki}, \href{https://tatoeba.org/en/}{Tatoeba}, \href{http://nlp.ffzg.hr/resources/corpora/setimes/}{Settimes}& Classifier+hierarch. softmax, $n$-grams& \\
Franc &$88/403$ & UDHR & $N$-grams \\
\hdashline
\ourLID & $517/517$  &  Several GDoc, SDoc, RDoc, News, ENC, IC  & Transformer \\
\bottomrule
\end{tabular}%
}
\caption{\ourLID~in comparison. \textbf{Covered/All:} \# of African lgs compared with covered lgs, \textbf{GDoc:} Gov docs, \textbf{SDoc:} Software docs, \textbf{RDoc:}Religious docs, \textbf{News:} Newswire, \textbf{ENC:} online encyclopedia, \textbf{IC:} Internet crawl. }
\label{tab:tools_comparison}
\end{table*}
\section{Diagnostic Case Studies}\label{sec:twitter}
Although \ourLID~is not trained on Twitter data, we evaluate its performance on Twitter to investigate the robustness of our models in out of domain scenarios. Namely, we carry out two diagnostic case studies using Twitter data. In the first study, which we refer to as Twitter in the wild, we use unannotated Tweets crawled from the web. In the the second, we use annotated tweets. We now turn to the details of these studies. 
\subsection{Case Study I: AfroLID in the Wild}\label{sec:results}
In order to evaluate the utility of \ourLID~in a real-world scenario, we collect $700$M tweets from Africa. For this, we use Twitter streaming API from $2021-2022$ with four geographical bounding boxes (central, eastern, western, and southern of Africa). We extract a random sample of $1$M tweets from this larger Twitter dataset for our analysis. As is known, Twitter currently automatically labels a total of $65$ languages. Only one of these languages, i.e., Amharic, is an African language in our $517$ languages. In the $1$M sample, $110$ tweets were tagged as "Amharic" and $6,940$ as "undefined" by Twitter. We run our model on the "undefined" data. In all, the $6,940$ tweets were identified as belonging to $242$ African languages by \ourLID. Since the Tweets we used were unannotated, we are not able to determine the number of tweets wrongly classified by~\ourLID~for each language. For this reason, we only evaluate a subset of the predicted languages: we ask native speakers of three languages (Y{o}r\`{u}b\'{a}, Hausa, and Nigerian Pidgin) to help identify each tweet that was classified by \ourLID~as belonging to their language. We provide details of this annotation study and examples of annotated samples in Table~\ref{tab:tweeter_in_d_wild} ( Appendix~\ref{sec:apx_twitter_lit}). We find that \ourLID~is able to correctly identify Y{o}r\`{u}b\'{a} both with and without diacritics and code-mixed examples. A total of $16$ tweets are classified as Y{o}r\`{u}b\'{a} by \ourLID, of which $7$ are correct ($43.75\%$), $2$ are mixed with English, and $7$ are wrongly labelled. Of the wrongly labelled tweets, one is identified as Nigerian Pidgin, while the others are unknown languages. For Nigerian Pidgin, of the $28$ tweets predicted, $2$ are correct ($12.50\%$), $1$ is mixed with an unknown language, and the others are wrongly classified. We find that in most cases, tweets classified as Nigerian pidgin are code-mixed with English and another Indigenous language. This gives us indication that \ourLID~identifies Nigerian Pidgin as an English-based Creole. Finally, a total of $333$ tweets are classified as Hausa. Of these, $105$ examples are correct ($37.50\%$), $18$ are mixed, while the others are wrongly labeled. 


\subsection{Case Study II: AfroLID on AfriSenti}
\begin{figure}[h!]
  \centering
 \includegraphics[width=\linewidth]{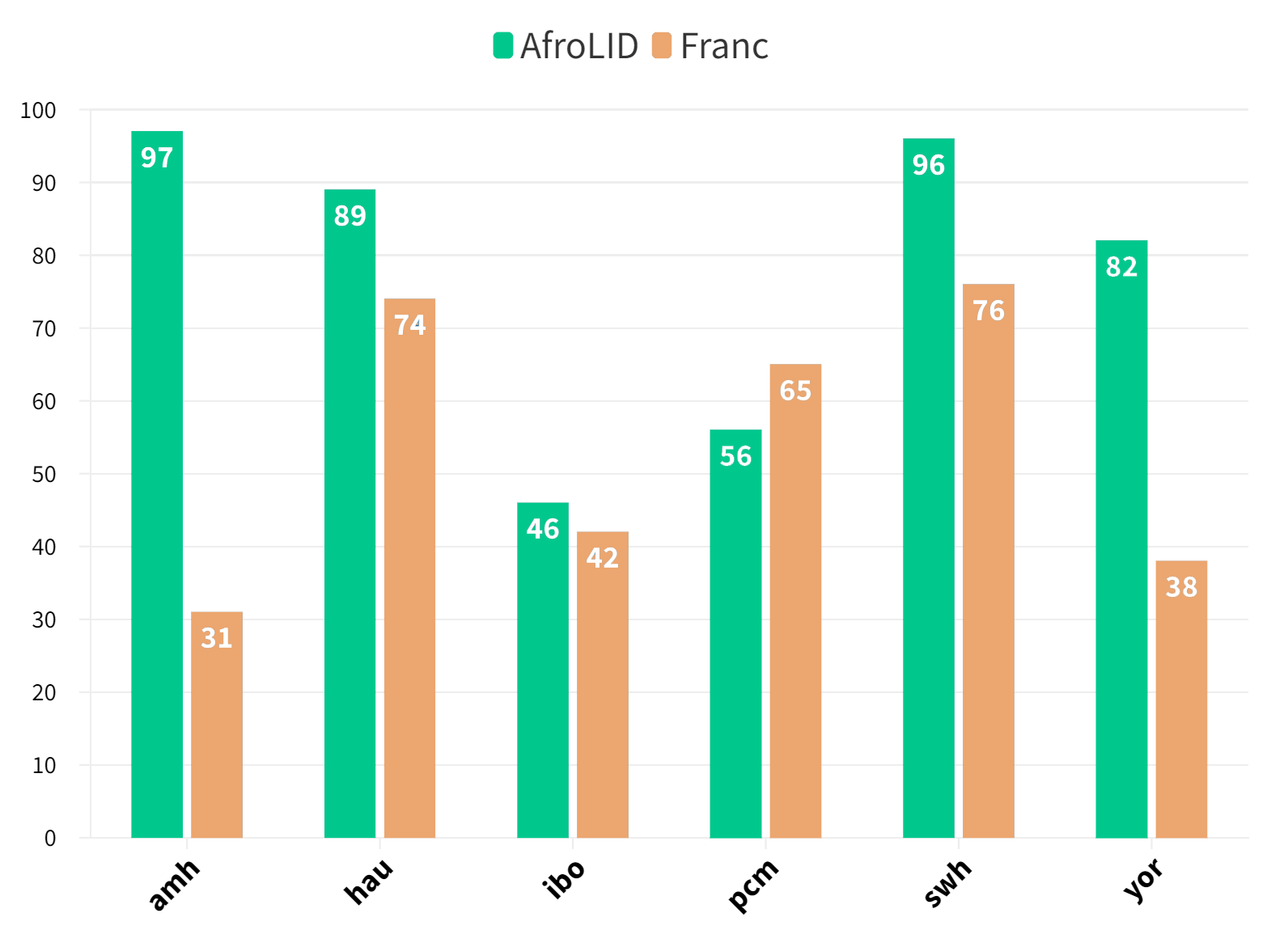}
\caption{\small Performance of \ourLID~and Franc on Afri-senti using $F_1$-score.}\label{fig:naija_senti} 
\end{figure}

We also test performance of \ourLID~on the recently released AfriSenti Twitter dataset of African languages. AfriSenti~\cite{muhammad2022naijasenti, yimam-etal-2020-exploring} contains ${\sim56,000}$ tweets annotated for sentiment in Amharic, Hausa, Igbo, Nigerian Pidgin, Swahili, and Y{o}r\`{u}b\'{a}. We run \ourLID~and Franc tool on AfriSenti. As Figure~\ref{fig:naija_senti} shows, \ourLID~outperforms Franc on all languages except Nigerian Pidgin. We assume this is because Franc supports English and may have learnt some lexical / grammatical information from English to aid the identification of Nigerian Pidgin (although \ourLID~ outperforms Franc on Nigerian Pidgin on our Dev and Test as shown in Table \ref{tab:franc_vs_afrolid} and \ref{tab:franc_vs_afrolid_test}.


\section{Related Work}\label{sec:litreview}
LID tools are often used to select data to pre-train language models~\cite{buck-etal-2014-n} and, more generally, develop multilingual corpora~\cite{Buck-commoncrawl, Dunn2020, Scannell2007TheCP, OrtizSuarezSagotRomary2019}. For many languages, including African languages, LID tools are either not available or perform poorly~\cite{caswell2021quality,caswell-etal-2020-language}. A few works, however, have already focused on African language identification. For example,~\newcite{lid_asubiaro} cover Y{o}r\`{u}b\'{a}, Hausa, and Igbo. Similarly,~\newcite{8261150, 10.1007/978-3-030-34058-2_26} treat $10$ Indigenous South African official languages. In addition, a handful of other African languages are covered in LID tools such as CLD2 \cite{mccandless2010accuracy}, CLD3~\cite{salcianu2018compact}, Equilid~\cite{jurgens2017incorporating}, FastText, Franc, LangDetect~\cite{nakatani2010langdetect} and Langid.py~\cite{lui-baldwin-2012-langid} and works such as~\newcite{mageed2020microdialects,mageed2020marbert} and~\newcite{nagoudi-etal-2022-arat5}. We provide an extended literature review of language identification, related tools, as well as data and methods employed in Appendix~\ref{app:appendix_lit}. We also provide a comparison between available LID tools in terms of training data, methodology, and number of covered African languages in Table~\ref{tab:tools_comparison}. To the best of our knowledge, \ourLID~is the first publicly available LID tool covering a large number of African languages and varieties (n=$517$).

\section{Conclusion}\label{sec:conclusion} 
We introduced our novel African language identification tool, \ourLID. To the best of our knowledge, \ourLID~is the first publicly available tool that covers a large number of African languages and language varieties. \ourLID~also has the advantages of wide geographical coverage ($50$ African countries) and linguistic diversity. We demonstrated the utility of \ourLID~on non-Latin scripts, Creoles, and languages with close geographical proximity. We also empirically showed \ourLID's superiority to five available tools, including in performance in the wild as applied to the much-needed Twitter domain. In the future, we plan to extend \ourLID~to cover the top $100$ most popular languages of the world as well as code-switched texts.
\section{Limitations}\label{sec:limitations}
We can identify a number of limitations for our work, as follows:

\begin{itemize}
    \item \ourLID~does not cover high-resource, popular languages that are in wide use by large populations. This makes it insufficient as a stand-alone tool in real-world scenarios where many languages are used side-by-side. Extending \ourLID~to more languages, however, should be straightforward since training data is available. Indeed, it is our plan to develop \ourLID~in this direction in the future.
    \item \ourLID~recognizes only Indigenous African languages in monolingual settings. This limits our tool's utility in code-mixed scenarios, (although Creoles are like code-mixed languages). This is undesirable especially because many African languages are commonly code-mixed with foreign languages due to historical reasons~\cite{adebara-abdul-mageed-2022-towards}. Again, to improve accuracy in the future, it would be beneficial to add foreign languages support in code-mixed settings such as with English, French, and Portuguese. 
    \item Although we strive to test \ourLID~in real-world scenarios, we were not able to identify native speakers except from a small number of languages. In the future, we plan to work more with the community to enable wider analyses of our predictions. 
    
\end{itemize}

\section{Ethical Considerations}\label{sec:ethics}
Although LID tools are useful for a wide range of applications, they can also be misused. We release \ourLID~hoping that it will be beneficial to wide audiences such as to native speakers in need of better services like health and education. Our tool is also developed using publicly available datasets that may carry biases. Although we strive to perform analyses and diagnostic case studies to probe performance of our models, our investigations are by no means comprehensive nor guarantee absence of bias in the data. In particular, we do not have access to native speakers of most of the languages covered in \ourLID. This hinders our ability to investigate samples from each (or at least the majority) of the languages. We hope that future users of the tool will be able to make further investigations to uncover \ourLID's utility in wide real-world situations.     

\section*{Acknowledgements}\label{sec:acknow}
We gratefully acknowledge support from Canada Research Chairs (CRC), the Natural Sciences and Engineering Research Council of Canada (NSERC; RGPIN-2018-04267), the Social Sciences and Humanities Research Council of Canada (SSHRC; 435-2018-0576; 895-2020-1004; 895-2021-1008), Canadian Foundation for Innovation (CFI; 37771), Digital Research Alliance of Canada,\footnote{\href{https://alliancecan.ca}{https://alliancecan.ca}} UBC ARC-Sockeye,\footnote{\href{https://arc.ubc.ca/ubc-arc-sockeye}{https://arc.ubc.ca/ubc-arc-sockeye}}, Advanced Micro Devices, Inc. (AMD), and Google. Any opinions, conclusions or recommendations expressed in this material are those of the author(s) and do not necessarily reflect the views of CRC, NSERC, SSHRC, CFI, CC, AMD, Google, or UBC ARC-Sockeye. 

\bibliography{anthology}
\bibliographystyle{acl_natbib}
\appendix
\clearpage
\appendixpage
\addappheadtotoc
\counterwithin{figure}{section}
\counterwithin{table}{section}





\section{Results of \ourLID~on Dev Set}\label{subsec:appendix_results}
We report results from comparing \ourLID~ with CLD2, CLD3, Langid.py, LangDetect, and Franc on our Dev set in Table \ref{tab:res}.

\begin{table}[H]
\centering
\resizebox{\columnwidth}{!}{%
\begin{tabular}{lcccccc}
\toprule
\textbf{Lang.}  & \textbf{CLD2} & \textbf{CLD3} & \textbf{Langid.py} & \textbf{LangDetect} & \textbf{Franc} & \textbf{\ourLID}\\ 

\midrule
        afr & \textbf{$94.11$} & $88.23$ & $70.58$ & $92.15$ & $84.31$ & \textbf{$\bf94.11$} \\ 
        amh & - & $98.03$ & \textbf{$\bf100.00$} & - & $25.49$ & $98.03$ \\ 
        hau & - & $86.27$ & - & - & $82.35$ & \textbf{$\bf94.11$} \\ 
        ibo & - & $92.15$ & - & - & $90.19$ & \textbf{$\bf94.11$} \\ 
        kin &\textbf{$\bf88.23$} & - & $56.86$ & - & $52.94$ & $80.39$ \\ 
        lug & $74.50$ & - & - & - & $52.94$ & \textbf{$\bf86.27$} \\ 
        mlg & - & \textbf{$\bf98.03$} & $92.15$ & - & - & $96.07$ \\ 
        nya & - & \textbf{$\bf96.07$} & ~ & - & $82.35$ & \textbf{$\bf96.07$} \\ 
        sna & - & $86.27$ & - & - & $80.39$ & \textbf{$\bf96.07$} \\ 
        som & - & $96.07$ & - & - & $96.07$ & \textbf{$\bf98.03$} \\ 
        sot & - & \textbf{$\bf90.19$} & - & - & \textbf{$\bf90.19$} & $76.47$ \\ 
        swa & $92.15$ & $90.19$ & $86.27$ & \textbf{$\bf96.07$} & - & $92.15$ \\ 
        swc & $90.19$ & $96.07$ & \textbf{$\bf98.03$ } & \textbf{$\bf98.03$} & - & $74.50$ \\ 
        swh & $88.23$ & \textbf{$\bf96.07$} & $90.19$ & $90.19$ & $72.54$ & $74.50$ \\ 
        xho & - & $90.19$ & \textbf{$\bf94.11$} & - & $64.70$ & $82.35$ \\ 
        yor & - & $50.82$ & - & - & $39.21$ & \textbf{$\bf100.00$} \\ 
        zul & - & \textbf{$\bf86.27$} & $ $ & - & $37.25$ & $58.82$ \\ 
\bottomrule
\end{tabular}%
}
\caption{A comparison of results on \ourLID~with CLD2, CLD3, Langid.py, LangDetect, and Franc using $F_1$-score on the Dev set. A dash (``$-$") indicates that the tool does not support the language.}
\label{tab:res}
\end{table}
\section{Analysis of \ourLID~}\label{subsec:appendix_analysis}
\begin{figure}[h!]
  \centering
  \includegraphics[width=\linewidth]{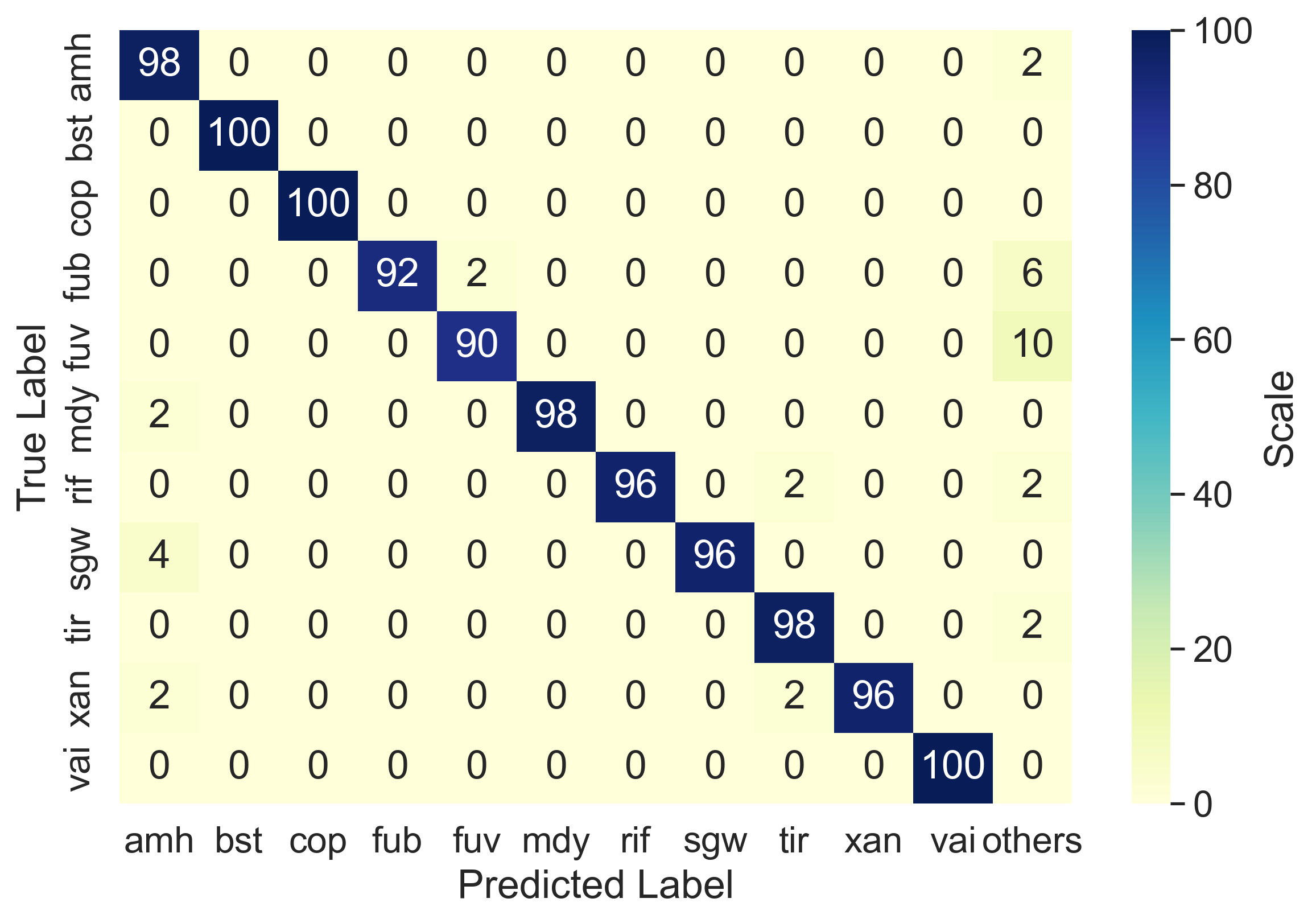}
\caption{\small Errors on the different script in \ourLID~Dev set. We use ISO-3 codes to represent the languages. "Others' refers to languages \ourLID~identifies as outside the list of languages selected for analysis.}\label{fig:s_confusion_matrix}
\end{figure}

\begin{figure}[h!]
  \centering
  \includegraphics[width=\linewidth]{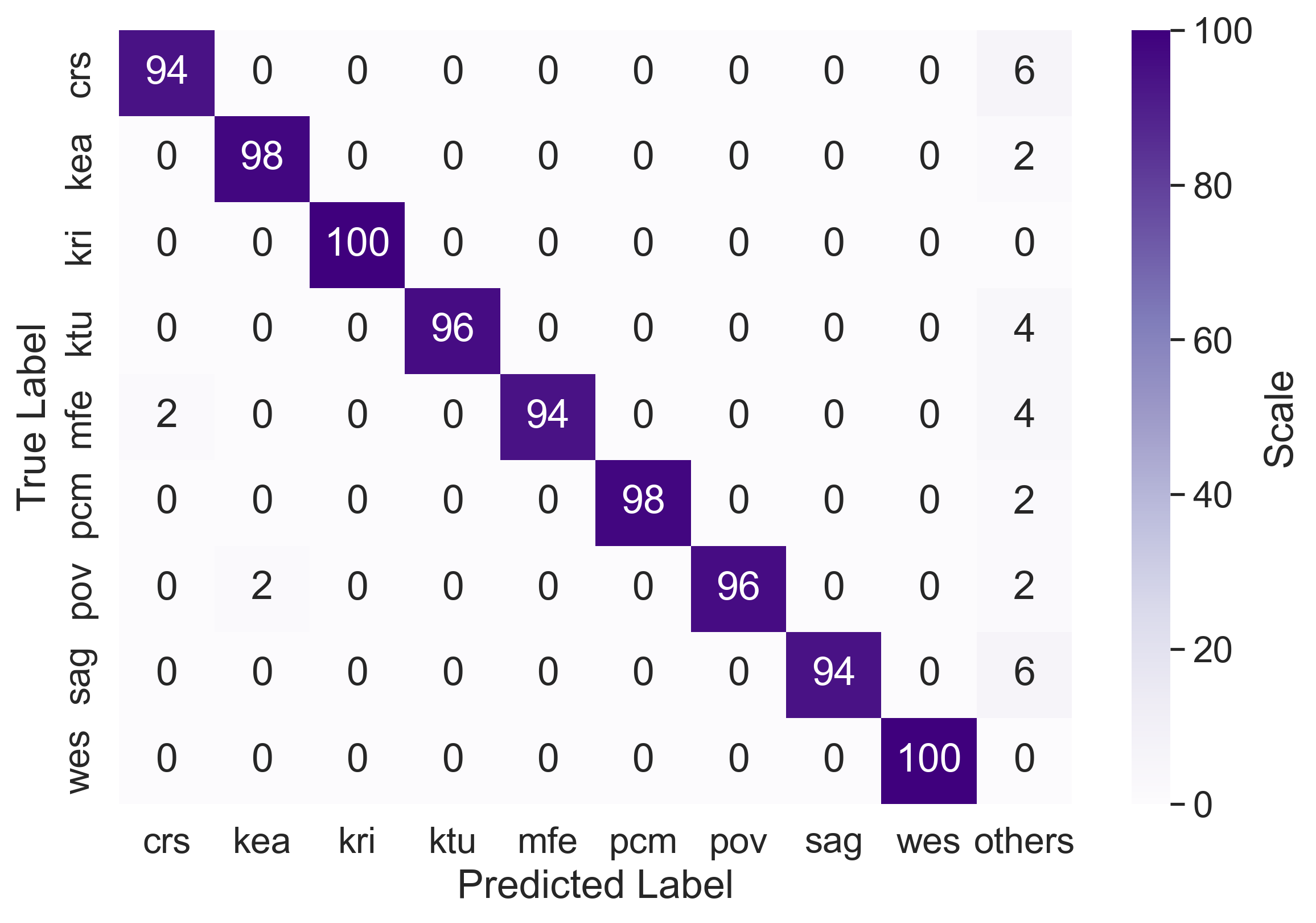}
\caption{\small Errors on the different Creoles in \ourLID. We use ISO-3 codes to represent the languages. ``Others" refers to languages \ourLID~identifies as outside the list of languages selected for analysis.}\label{fig:creole_conf_matrix} 
\end{figure}

\begin{figure}[h!]
  \centering
  \includegraphics[width=\linewidth]{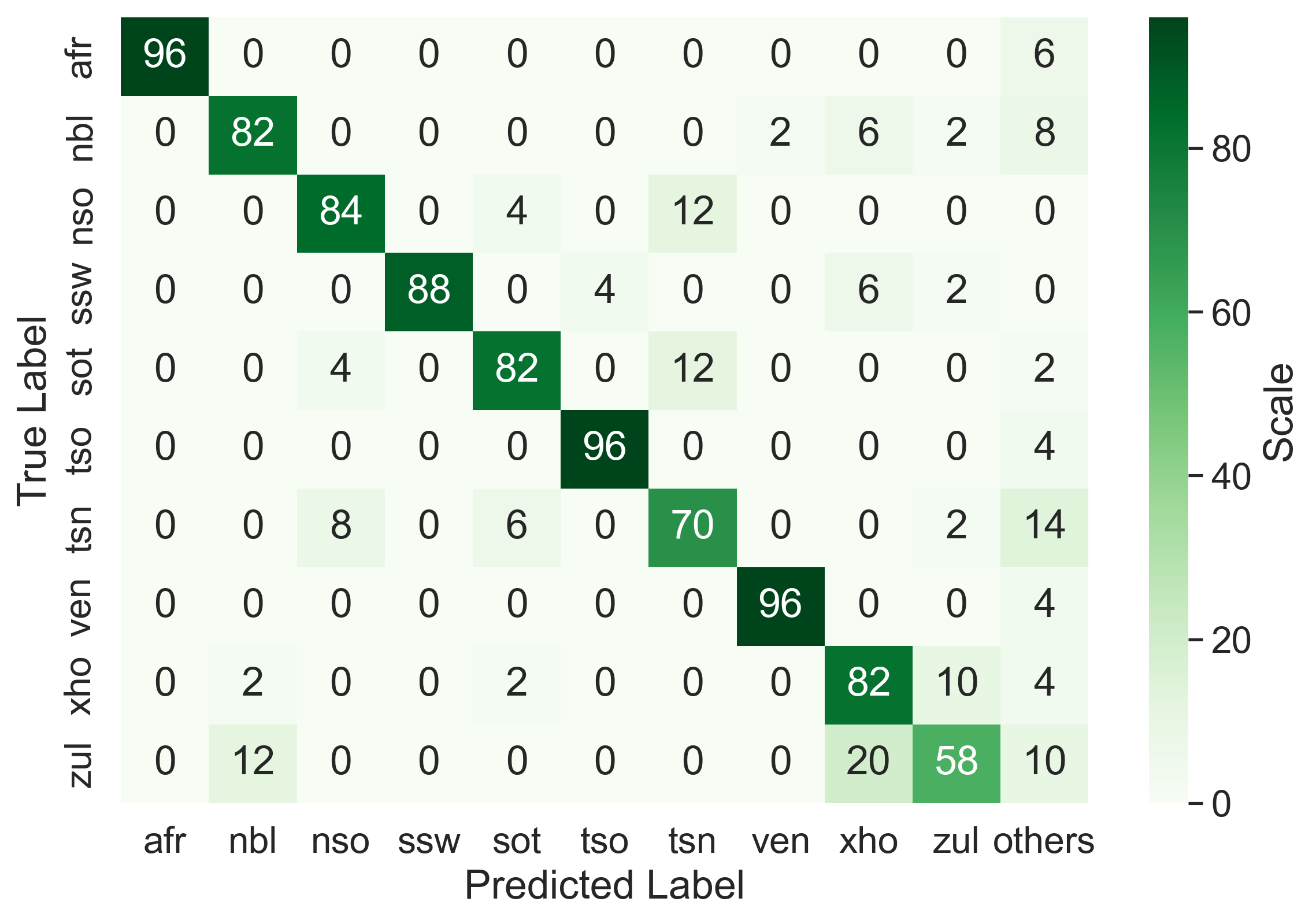}
\caption{\small Errors on Indigenous South African languages in \ourLID~Dev data. "Others' refers to languages \ourLID~identifies as outside the list of languages selected for analysis.}\label{fig:sa_conf_matrix}

\end{figure}

We perform the experiments on non-Latin scripts, Creoles, and languages in close geographical proximity on the Dev set, as in Subsection \ref{sec:perform_analysis}. We show the results on the performance of \ourLID~on non-Latin scripts in Table \ref{fig:s_confusion_matrix}, Creole languages in Table \ref{fig:creole_conf_matrix} and geographical proximity in Table \ref{fig:sa_conf_matrix} respectively.

\section{Extended Literature Review}\label{app:appendix_lit}
\subsection{Datasets}
Datasets for LID are often created using various genre of data for one or more languages. For multilingual LID, which is the focus of our work, documents are gathered from web pages containing multiple languages. Web pages for multilingual organizations are also often desirable because the same text is translated into various languages. Most datasets for multilingual LID cover European languages and many other high resource languages, making \ourLID~dataset a significant contribution to AfricaNLP. To the best of our knowledge, \ourLID~dataset is the first publicly available dataset for multilingual language identification for African languages. We provide details of some other publicly available corpora for LID.

\textbf{DSL Corpus Collection} \cite{tan:2014:BUCC, malmasi-etal-2016-discriminating, zampieri-etal-2015-overview, zampieri-etal-2014-report} is a multilingual collection of short excerpts of journalistic texts. It has been used as the main data set for the DSL shared tasks organized within the scope of the workshop on NLP for Similar languages, Varieties and Dialects (VarDial). It covers 22 languages. 

\textbf{NLI-PT} \cite{del-rio-gayo-etal-2018-portuguese} is a dataset collected from three different learner corpora of Portuguese including COPLE2; Leiria corpus, and PEAPL. The three corpora contain written productions from learners of Portuguese with different proficiency levels and native languages. The dataset included all the data in COPLE2 and sections of PEAPL2 and Leiria corpus with details of the dataset in Table \ref{tab:NLI-PT}. Therefore, the dataset include texts corresponding to the following 15 languages:
Arabic, Chinese, Dutch, English, French, German, Italian, Japanese, Korean, Polish, Romanian, Russian, Swedish, Spanish, and Tetum.
\begin{table}[]
\centering
\resizebox{0.9\columnwidth}{!}{%
\begin{tabular}{lllll}
\toprule

 & {\textbf{COPLE2}} & {\textbf{LEIRIA}} & {\textbf{PEAPL2}} & {\textbf{TOTAL}} \\ \midrule
Sents & $1,058$ & $330$ & $480$ & $1,868$ \\ 
Tokens & $201,921$ & $57,358$ & $121,138$ & $380,417$ \\ 
Types & $9,373$ & $4,504$ & $6,808$ & $20,685$  \\ 
TTR & $0.05$ & $0.08$ & $0.06$ & $0.05$  \\ \bottomrule

\end{tabular}%
}
 \caption{ Distribution of the dataset: Number of texts,
tokens, types, and type/token ratio (TTER) per source
corpus.
} \label{tab:NLI-PT}
 \end{table}

\textbf{Wanca 2017 Web Corpora} \cite{jauhiainen-etal-2020-uralic} is made up of re-crawls performed by the SUKI project. The target of the re-crawl was to download and check the availability of the then current version of the Wanca service of about $106,000$ pages. This list of $106,000$ http addresses was the result of several earlier web-crawls, in which they had identified the language in a total of $3,753,672,009$ pages. 

\textbf{EUROGOV, TCL, and WIKIPEDIA} \cite{baldwin-lui-2010-language} consist of documents with a single encoding across $10$ European languages; shorter documents across different encodings for $60$ languages, and wikipedia web crawls for $67$ languages respectively. These collection cover different genres with Eurogov collected from government documents, TCL from online news sources and Wikipedia dumps.

\textbf{The UMass Global English on Twitter Dataset} \cite{blodgett-etal-2017-dataset} contains $10,502$ tweets, randomly sampled from all publicly available geotagged Twitter messages, annotated for being in English, non-English, or having code switching, language ambiguity, or having been automatically generated. It includes messages sent from $130$ different countries.

\subsection{Features}
Different features can be used for training a LID system including:
\begin{itemize}
    \item Bytes and Encoding: Some encodings use a fixed number of bytes e.g ASCII while some others use variable length encoding. Some languages also use specific encodings (GuoBiao 18030 or Big5 for chinese) while the same encoding can be used for different languages (e.g UTF-8).
    
    \item Characters: Non-alphabetic, alphabets, capitalization, the number of characters in words and word combinations, the number of characters in words and word combinations have been used as features. Non-alphabetic characters has been used to detect languages like Arabic, emojis, and other languages that use non-alphabetic characters \cite{samih_2017, bestgen-2017-improving, Dongen2017AnalysisAP}. Alphabets can also be used to exclude languages where a unique character is absent in the test document. 
    \item Character combination: co-occurrences of some characters can be used to detect some languages. Linguistically, some languages abhor certain combination of characters which some other languages allow. For example some Niger-Congo languages abhor vowel hiatus and every consonant must be followed by a vowel. This feature has been found useful for developing LID systems \cite{van-der-lee-van-den-bosch-2017-exploring, Dongen2017AnalysisAP, Martinc2017PAN2A}. 
    
    \item  Morphemes, Syllables and Chunks: different morphological features including prefixes, suffixes, and character n-grams \cite{gomez-etal-2017-discriminating}. Syllables, chunks, and chunks of syllables / ngrams have also been used for LID. This also has linguistic significance in that the prefix, suffixes and morphological information embedded in a language can provide information about the etymology of a language. 
    
    \item Words: The position of words \cite{adouane-dobnik-2017-identification}, the string edit distance and n-gram overlap between the word to be identified and words in dictionaries, dictionary of unique words in a language, basic dictionary of a language, most common words, word clusters among others are some discriminating features used for LID.
    
    \item Combination of words: Here, length of words, the ratio to the total number of words of: once-occurring words, twice-occurring words, short words, long words, function words, adjectives and adverbs, personal pronouns, and question words are some features used here \cite{van-der-lee-van-den-bosch-2017-exploring}. This feature is linguistically significant since the ratio of certain categories of words can be useful for identifying some languages. 
    
    \item Syntax and Part of speech (POS) tags: Syntactic features can be used to identify languages. Identifying an adjective before a noun for instance may be a good indication for some languages and even the tags available can be a useful feature. Syntactic parsers together with dictionaries and morpheme lexicons, n-grams composed of POS tags and function words have all been used as features \cite{adouane-dobnik-2017-identification} for LID. 
    
    \item Languages identified for surrounding words in word-level LID: The language of surrounding words can also be a useful feature since there may be a higher likelihood of having some languages used together. This is especially true in the case of codeswitching where some languages are more likely to be used together than some others \cite{Dongen2017AnalysisAP}. 
    
    \item Feature smoothing: Feature smoothing is required in order to handle the cases where not all features in a test document have been attested in the training corpora. Feature smoothing is used in low resource scenarios and when the frequency of some features are high. Different types of feature smoothing is possible. Some of them are additive smoothing where an extra number of occurrences is added to every possible feature in the language model \cite{Jauhiainen2019AutomaticLI}. 

\end{itemize}

\subsection{Methods}
Algorithms for LID work by first using one or more features before using a classification algorithm to determine the appropriate language for a text\cite{grothe-etal-2008-comparative, Jauhiainen2019AutomaticLI}. 

\textbf{Hidden Markov Models (HMM)}
Hidden Markov Models (HMM) are commonly used in spoken language identification \cite{ZISSMAN2001115, Yonghong_1995} as well as for written language \cite{guzman-etal-2016-simple}. Language models are first trained for each language that the system must know about using a text corpora, and stored for later comparison with unidentified text. In these models the parameters of the HMM are the transition probability and the initial probability. Probabilities are calculated using the relative frequency of each transition or initial state of the training data. After training, the system calculates the sequence probability using each language model that has been trained \cite{Padr2004ComparingMF}. 

\textbf{N-Gram-Based Text Categorization}
This method introduced by \cite{Cavnar94n-gram-basedtext, grothe-etal-2008-comparative} is based on comparing unique n-gram frequency profiles. These frequencies are sorted in decreasing order for all unique n-grams. N-gram profiles are created for each language to be trained with $n= 1$ to $5$. To classify a piece of text, the n-gram frequency for that text is built and compared to the n-gram profiles calculated during the training phase. This is done by computing the distance between the n-gram profiles of the text and that for each language model. The computation also penalizes the total score of the language for each missing n-gram. The language with the lowest score is selected as the identified language \cite{jauhiainen-etal-2017-evaluation, Padr2004ComparingMF}.

\textbf{LIGA}
This uses a graph-based n-gram approach called LIGA which was originally used for sentiment analysis \cite{Tromp_multilingualsentiment} and adopted for LID \cite{vogel2012robust}. The language models use the relative frequencies of character trigrams and those of 4-grams. To identify the language in a text, the relative frequency of each trigram and 4-gram found in a language model is added to the score of the language. The language with the highest score is selected as the language of the text.

\textbf{HELI Method}
The HeLI method \cite{Jauhiainen_2017} uses character n-grams based language models for each language. The n-gram values are hyperparameters from one to a specific  maximum number $N\textsubscript{max}$. The model then selects one language model when classifying the language of a text. The selection is based on the most applicable model to the specified text. The model then gradually backs off to a lower order n-gram if the n-gram with the $N\textsubscript{max}$ is not applied until an n-gram can be applied. The validation set is used during evaluation to determine the best values for $N\textsubscript{max}$, the maximum number of features to be included in the language models, and the penalty for languages without the selected feature. The penalty functions like a smoothing parameter by transferring some of the probability mass to unseen features in the language model \cite{jauhiainen-etal-2017-evaluation}. 

\textbf{Whatlang program}
This uses language models built with n-grams of variable byte lengths between $3-12$~\cite{Brown_2013}. The K most frequent n-grams and their relative frequencies are then extracted and calculated for each language. Once the first model is generated, substrings of larger n-grams are filtered out if the larger n-gram has a frequency not less than $62\%$ of the frequency of the shorter n-grams. The model weights are computed for each language such that shorter n-grams with the same relative frequency have lower weights than those with larger n-grams. This is because larger n-grams are more informative but less common.

\subsection{Language Identification Tools}
Several tools have been developed for multilingual LID. We provide details of different tools which has representation for African languages including CLD2~\cite{mccandless2010accuracy}, CLD3~\cite{salcianu2018compact} EquiLID~\cite{jurgens2017incorporating}, fastText \cite{joulin-etal-2017-bag}, \href{https://github.com/wooorm/franc}{Franc}, Langid.py \cite{lui-baldwin-2012-langid}, and LangDetect \cite{nakatani2010langdetect}. 

\subsubsection{CLD2\footnote{\href{https://github.com/CLD2Owners/cld2}{https://github.com/CLD2Owners/cld2}}}
CLD2~\cite{mccandless2010accuracy} covers $83$ languages and trained on web pages text, using one of three different token algorithms. CLD2 probabilistically detects over 86 languages including Afrikaans and Swahili. Unicode UTF-8 text, either plain text or HTML/XML. It requires that legacy encodings be converted to valid UTF-8. For mixed-language input, CLD2 returns the top three languages found and their approximate percentages of the total text bytes (e.g. $80\%$ English and $20\%$ French out of $1000$ bytes of text means about $800$ bytes of English and $200$ bytes of French). Optionally, it also returns a vector of text spans with each language identified. 

\subsubsection{CLD3}
CLD3~\cite{salcianu2018compact}\footnote{\href{https://github.com/google/cld3}{https://github.com/google/cld3}}, the latest updated version of \textit{CLD2} (2020) covers $106$ languages including Afrikaans, Amharic, Hausa, Malagasy, Shoma, Somali, Swahili, Xhosa, Yoruba, and Zulu. CLD3 uses a neural network model for language identification. It contains the inference code and a trained model. 

\subsubsection{EquiLID}
EquiLID~\cite{jurgens2017incorporating}\footnote{\href{https://github.com/davidjurgens/equilid}{https://github.com/davidjurgens/equilid}} is a character based DNN $encoder-decoder$ model~\cite{cho-etal-2014-learning, sutskever2014sequence} with an attention mechanism~\cite{bahdanau2014neural}. Equilid is a general purpose language identification library and command line utility built to identify a broad coverage of languages, recognize language in social media, with a particular emphasis on short text, recognizing dialectic speech from a language's speakers, identify code-switched text in any language pairing at least at the phrase level, provide whole message and per-word. EquiLID covers $70$ languages including Amharic. 

\subsubsection{FastText}
FastText \cite{joulin2016fasttext} supports $176$ languages including $5$ African languages. The model uses a classifier with hierachical softmax with n-grams. 

\subsubsection{Franc}
\href{https://github.com/wooorm/franc}{Franc} supports $403$ languages including $88$ African languages. It is built using Universal Declaration of Human Rights \href{http://unicode.org/udhr/}{UDHR} documents translated into multiple languages. Details of the model architecture is not available, however there is indication that $n$-grams are used in the model.

\subsubsection{LangDetect}
LangDetect \cite{nakatani2010langdetect} covers $49$ languages including Afrikaans and Swahili. LangDetect uses a huge dictionary of inflections and compound words over a Naive Bayes model with character n-grams.

\subsubsection{Langid.py}
\texttt{Langid.py} \cite{lui-baldwin-2012-langid} covers $97$ languages including Afrikaans, Amharic, Malagasy, Kinyarwanda, Swahili, and Zulu. The model is trained over a naive Bayes classifier with a multinomial event model using a mixture of byte n-grams. \texttt{langid.py} was designed to be used off-the-shelf. It comes with an embedded model using training data drawn from $5$ domains - government documents, software documentation, newswire, online encyclopedia, and an internet crawl, though no domain covers the full set of languages by itself, and some languages are present only in a single domain. Different aspects of \texttt{langid.py} are evaluated in different ways. For cross-lingual feature selection evaluation, each dataset is partitioned into two sets of equal sizes. The first partition is used for training a classifier while the second is used for evaluation. Since each dataset covers a different set of languages, there may be languages in the evaluation dataset that are not present in the training dataset \cite{lui-baldwin-2011-cross}. The \texttt{langid.py} module on the other hand is evaluated on different datasets and the accuracy is compared with those for CLD, Textcat, and LangDetect. The accuracy of Langid.py exceeded those from other tools on two twitter datasets \cite{lui-baldwin-2012-langid}. \texttt{Langid.py} can be used as a command line tool, python library, or web service tool.

\begin{table}[!ht]
\begin{center}
\small 

\begin{tabular}{>{}c|l l}
 \toprule
\multicolumn{1}{c}{}  &\textbf{LID Tool} & \textbf{African Languages } \\ 
 \midrule
\multicolumn{1}{c}{}  & CLD2 & afr, lug, kin, swa \\
\multicolumn{1}{c}{}  & CLD3 & afr, amh, hau, ibo, mlg, nya, sna,\\
\multicolumn{1}{c}{}  & & som, sot, swa, xho, yor, zul \\
\multicolumn{1}{c}{}  & Langid.py & afr, amh, kin, mlg, swa, xho, zul \\
\multicolumn{1}{c}{}  & EquiLID & amh  \\
\multicolumn{1}{c}{}  & LangDetect & afr, swh \\
\multicolumn{1}{c}{}  & FastText & afr, amh, mlg, som, swh, yor \\
 \bottomrule
\end{tabular}
\end{center}
\caption{African languages represented in different LID tools.}\label{tab:lid_tools}
\end{table}

Other LID tools without representation of African languages include
\href{https://github.com/shuyo/ldig}{LDIG}, and 
\href{https://github.com/microsoft/LID-tool}{Microsoft LID-tool} \cite{Gella2013QueryWL, gella-etal-2014-ye} which is a word level language identification tool for identifying code-mixed text of languages (like Hindi etc.) written in roman script and mixed with English.

\begin{table*}[h!]
\begin{tabular}{l l l l l l l l l l l l }
\toprule
{ aar}                                                        & { bez}                                                        & { cou}                                                        & { eza}                                                        & { ife}                                                        & { khy}                                                        & { lem}                                                        & { mfi}                                                        & { nga}                                                        & { rif}                                                        & { ssc}                                                        & { { uth}}                                                  \\
{ abn}                                                        & { bfa}                                                        & { csk}                                                        & { { fia}}                                                  & { igb}                                                        & { kia}                                                        & { lik}                                                        & { mgc}                                                        & { ngb}                                                        & { rim}                                                        & { suk}                                                        & { vag}                                                        \\
{ {ada}}                                                  & { bfd}                                                        & { daa}                                                        & { fip}                                                        & { ige}                                                        & { kik}                                                        & { lip}                                                        & { mgo}                                                        & { ngn}                                                        & { rub}                                                        & { sus}                                                        & { vif}                                                        \\
{ {adj}}                                                  & { bfo}                                                        & { daf}                                                        & { flr}                                                        & { igl}                                                        & { kkj}                                                        & { lmd}                                                        & { mgq}                                                        & { nhr}                                                        & { run}                                                        & { taq}                                                        & { vun}                                                        \\
{ {afr}}                                                  & { bib}                                                        & { dga}                                                        & { fon}                                                        & { ijn}                                                        & { klu}                                                        & { lmp}                                                        & { mkl}                                                        & { nhu}                                                        & { rwk}                                                        & { tcd}                                                        & { vut}                                                        \\
{ agq}                                                        & { biv}                                                        & { dgi}                                                        & { gaa}                                                        & { ikk}                                                        & { kmb}                                                        & { lnl}                                                        & { mlr}                                                        & { nim}                                                        & { sag}                                                        & { tem}                                                        & { wbi}                                                        \\
{ akp}                                                        & { bjv}                                                        & { dhm}                                                        & { gbo}                                                        & { ikw}                                                        & { knf}                                                        & { log}                                                        & { mnf}                                                        & { nin}                                                        & { sba}                                                        & { tex}                                                        & { wib}                                                        \\
{ ann}                                                        & { bky}                                                        & { dib}                                                        & { gid}                                                        & { iqw}                                                        & { koq}                                                        & { lol}                                                        & { mnk}                                                        & { niq}                                                        & { sbd}                                                        & { tgw}                                                        & { wmw}                                                        \\
{ anu}                                                        & { bmo}                                                        & { did}                                                        & { giz}                                                        & { iri}                                                        & { kqp}                                                        & { lom}                                                        & { mos}                                                        & { niy}                                                        & { sbp}                                                        & { thk}                                                        & { xed}                                                        \\
{ anv}                                                        & { bmv}                                                        & { dik}                                                        & { gkp}                                                        & { iso3}                                                       & { kqs}                                                        & { loq}                                                        & { moz}                                                        & { nko}                                                        & { sef}                                                        & { thv}                                                        & { xpe}                                                        \\
{ asg}                                                        & { bom}                                                        & { dip}                                                        & { gna}                                                        & { izr}                                                        & { krs}                                                        & { lot}                                                        & { mpg}                                                        & { nla}                                                        & { ses}                                                        & { tiv}                                                        & { xrb}                                                        \\
{ atg}                                                        & { bov}                                                        & { dnj}                                                        & { gnd}                                                        & { izz}                                                        & { krw}                                                        & { loz}                                                        & { mqb}                                                        & { nnh}                                                        & { sev}                                                        & { tlj}                                                        & { xsm}                                                        \\
{avn}                                                        & { box}                                                        & { dow}                                                        & { gng}                                                        & { jgo}                                                        & { krx}                                                        & { lro}                                                        & { mua}                                                        & { nnw}                                                        & { sfw}                                                        & { tod}                                                        & { xtc}                                                        \\
{ {avu}}                                                  & { bqc}                                                        & { dsh}                                                        & { gol}                                                        & { jib}                                                        & { ksb}                                                        & { luc}                                                        & { muh}                                                        & { nse}                                                        & { shi}                                                        & { tog}                                                        & { xuo}                                                        \\
{ azo}                                                        & { bqj}                                                        & { dug}                                                        & { gqr}                                                        & { kam}                                                        & { ksf}                                                        & { lwo}                                                        & { muy}                                                        & { nso}                                                        & { shj}                                                        & { tsw}                                                        & { yam}                                                        \\
{ bav}                                                        & { bsc}                                                        & { dyi}                                                        & { gso}                                                        & { kbn}                                                        & { ksp}                                                        & { maf}                                                        & { mwm}                                                        & { nus}                                                        & { shk}                                                        & { ttq}                                                        & { yao}                                                        \\
{ bba}                                                        & { bss}                                                        & { ebr}                                                        & { gur}                                                        & { kbo}                                                        & { kss}                                                        & { mbu}                                                        & { mws}                                                        & { nyb}                                                        & { sig}                                                        & { ttr}                                                        & { yat}                                                        \\
{ bbj}                                                        & { bud}                                                        & { ebu}                                                        & { guw}                                                        & { kbp}                                                        & { kub}                                                        & { mcp}                                                        & { myb}                                                        & { nyy}                                                        & { sil}                                                        & { tui}                                                        & { yba}                                                        \\
{ bbk}                                                        & { bum}                                                        & { efi}                                                        & { gux}                                                        & { kcg}                                                        & { kuj}                                                        & { mcu}                                                        & { myk}                                                        & { nza}                                                        & { snf}                                                        & { tul}                                                        & { yor}                                                        \\
{ bci}                                                        & { bus}                                                        & { ego}                                                        & { gvl}                                                        & { kde}                                                        & { kyq}                                                        & { mda}                                                        & { mzm}                                                        & { odu}                                                        & { snw}                                                        & { tum}                                                        & { zga}                                                        \\
{ bcp}                                                        & { buy}                                                        & { eka}                                                        & { gya}                                                        & { kde}                                                        & { kzr}                                                        & { mdm}                                                        & { mzw}                                                        & { okr}                                                        & { sop}                                                        & { tvu}                                                        & { zne}                                                        \\
{ bcy}                                                        & { bza}                                                        & { etu}                                                        & { hna}                                                        & { kdh}                                                        & { lam}                                                        & { meq}                                                        & { naq}                                                        & { oku}                                                        & { sor}                                                        & { udu}                                                        & { }                                                           \\
{ bdh}                                                        & { bzw}                                                        & { etx}                                                        & { ibb}                                                        & { kdl}                                                        & { lap}                                                        & { mer}                                                        & { ncu}                                                        & { ozm}                                                        & { sot}                                                        & { umb}                                                        & { }                                                           \\
{ bds}                                                        & { cko}                                                        & { ewe}                                                        & { ibo}                                                        & { ken}                                                        & { lee}                                                        & { mev}                                                        & { ndv}                                                        & { pkb}                                                        & { soy}                                                        & { urh}                                                        & { }                                                           \\
{ bex}                                                        & { cme}                                                        & { ewo}                                                        & { idu}                                                        & { ker}                                                        & { lef}                                                        & { mfh}                                                        & { ndz}                                                        & { pko}                                                        & { spp}                                                        & { uth}                                                        & { }                                                      \\
\toprule
\end{tabular}
\caption{Language varieties that use diacritics in our training data.}
\label{tab:diacritics_in_training}
\end{table*}

\section{Twitter Analysis}\label{sec:apx_twitter_lit}
For the Twitter in the wild analysis, we ask for annotations of \textit{yes}, \textit{no} or \textit{mixed} on each tweet, where \textit{yes} indicates agreement with the predicted label, \textit{no} indicates disagreement, and \textit{mixed} indicates that the tweet contains one or more other language than the predicted. We also ask for further annotations if the tweet is not in the predicted language, or is mixed with another/other language(s). In these cases, respondents are asked to identify the correct language (or mixed language[s]) if they know the language(s). We provide example annotation in the wild analysis in Table~\ref{tab:tweeter_in_d_wild} \begin{table*}[h!]
\resizebox{2\columnwidth}{!}{%
\begin{tabular}{
>{}l 
>{}l 
>{}l 
>{}l 
>{}l }
\toprule
\multicolumn{1}{c}{{\textbf{ISO-3}}} & \multicolumn{1}{l}{{\textbf{Tweet}}}                           & {\textbf{Representative?}}                     & {\textbf{No}}                                  & {\textbf{Mixed}}                               \\ \midrule
{}                                   & {Don't be on my TL supporting a rapist, a o ní s'oriburubuku o}                       & {Mixed}                                        & {}                                             & {English}                                      \\
{}                                   & {USER Omo ilorin Nile Adeleke ti Binu}                                           & {Yes}                                          & {}                                             & {}                                             \\
{}                                   & {Oproblema opo openi ne}                                                            & {No}                                           & {Unknown}                                      & {}                                             \\
\multirow{-4}{*}{{yor}}              & {USER On top Iron Konji na Bastard}                                              & {No}                                           & {Nigerian Pidgin}                              & {}                                             \\ \hline
{}                                   & {USER Mana ima na ife any i na-ekwu bu eziokwu}                                         & {Yes}                                          & {}                                             & {}                                             \\
{}                                   & {USER Mo je ri e}                                                                   & {No}                                           & {Y{o}r\`{u}b\'{a}}                                       & {}                                             \\
\multirow{-4}{*}{{ibo}}              & {USER Hamna namna mzee}                                                             & {No}                                           & {Unknown}                                      & {}                                             \\ \hline
{}                                   & {USER Kaji dadinka brother ka huta}                                                    & {Mixed}                                        & {}                                             & {English}                                      \\
{}                                   & {USER Su Umar danbarade}                                                            & {Yes}                                          & {}                                             & {}                                             \\
{}                                   & {USER Good nkosazana Cathy}                                                       & {No}                                           & {English + unknown}                            & {}                                             \\
\multirow{-4}{*}{{hau}}              & {ovo ra mbuti USER Sesi Gladys mani}                                           & {No}                                           & {Unknown}                                      & {}                                             \\ \hline
{}                                   & {USER Gompieno o bone dust !}                                                       & {Mixed}                                        & {}                                             & {Unknown}                                      \\
{}                                   & {USER Wey I travel from Ilesa to Ipetumodu}                                            & {Yes}                                          & {}                                             & {}                                             \\
{}                                   & {USER Ende zwotoralo ngoho ngoho}                                                      & {No}                                           & {Unknown}                                      & {}                                             \\
\multirow{-4}{*}{{pcm}}              & {Despacito! beyaudkrnkwudh despacito, daueiejrb despacitoo! goose bumps} & {No}                                 & {English + unknown}                            & {}  \\\bottomrule                                      
\end{tabular}
}
\caption{Some example annotations for the Twitter in the wild analysis. We show for each language the 4 possible annotations.}\label{tab:tweeter_in_d_wild}
\end{table*}.

\section{ Languages Covered in \ourLID~}\label{sec:apx_covered_langs}
\begin{figure*}[ht]
  \centering
  \includegraphics[width=0.9\linewidth]{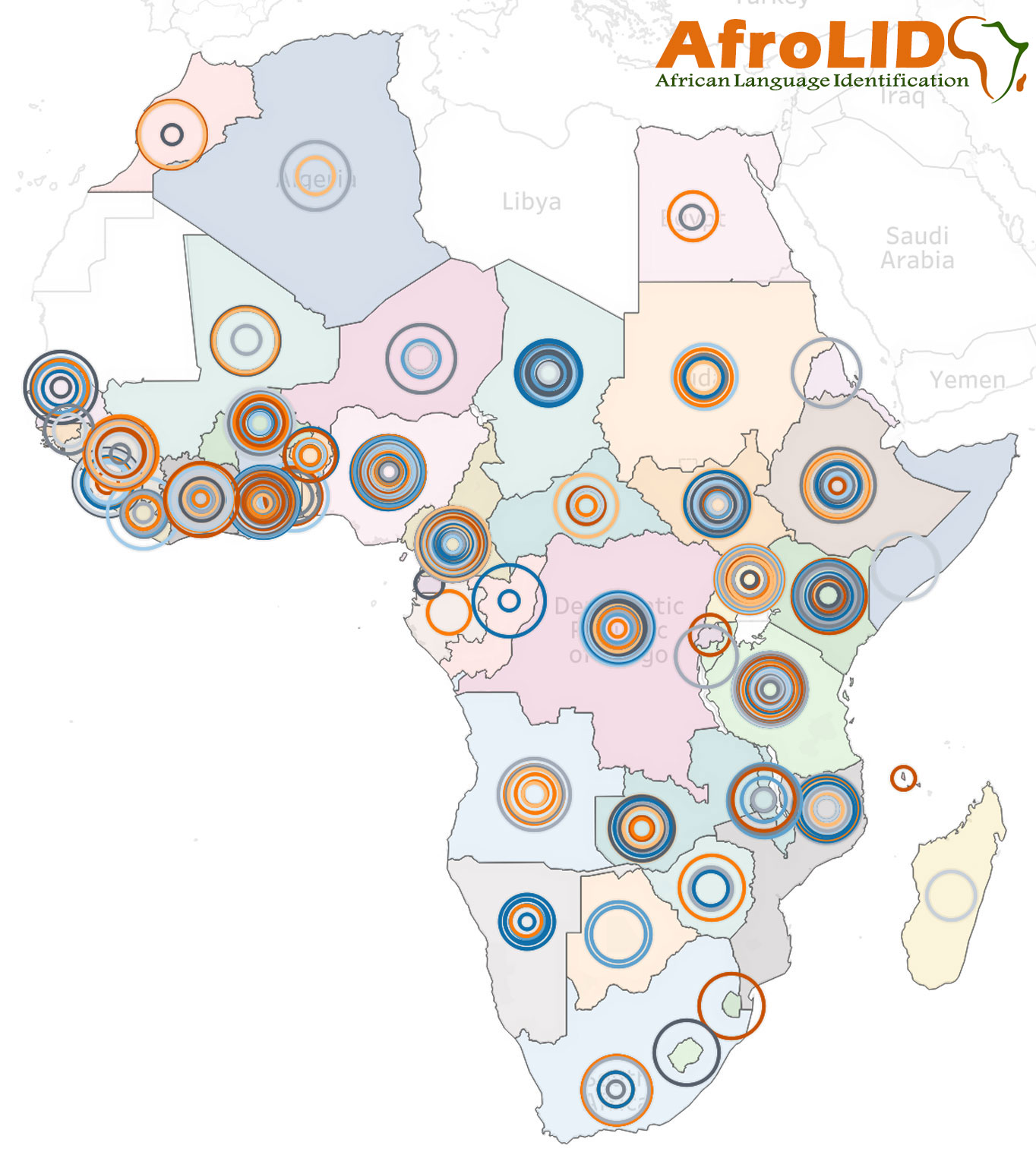}
\caption{\small All $50$ African countries in our data, with our $517$ languages/language varieties in colored circles overlayed within respective countries.}
\label{fig:map_lang_families} 
\end{figure*}

\begin{figure*}[ht]
  \centering
  \includegraphics[width=0.9\textwidth]{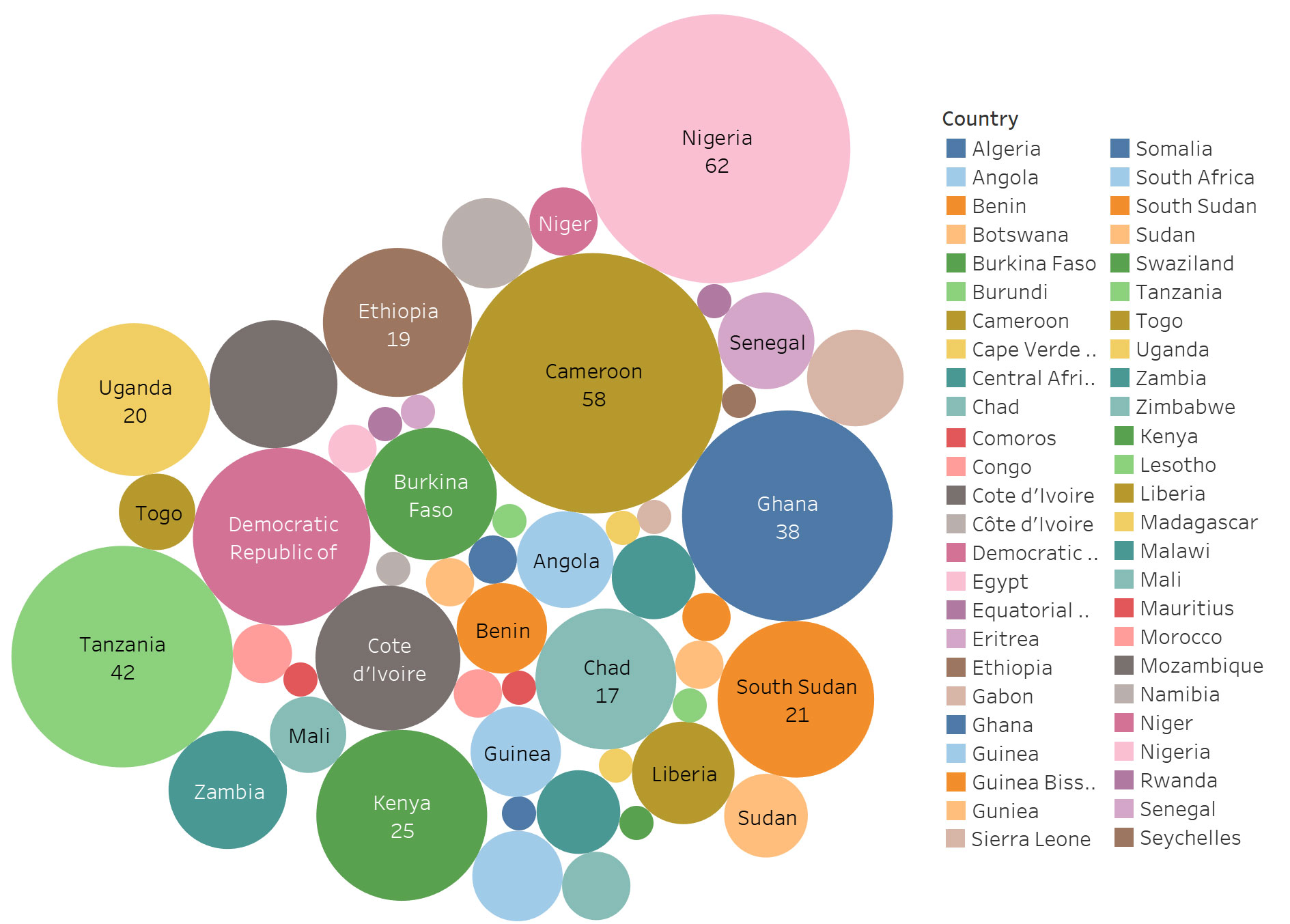}
\caption{\small \ourLID's Covered languages.}
\label{fig:apx_lang_families_A} 
\end{figure*}

\begin{figure*}[ht]
  \centering
  \includegraphics[width=0.9\linewidth]{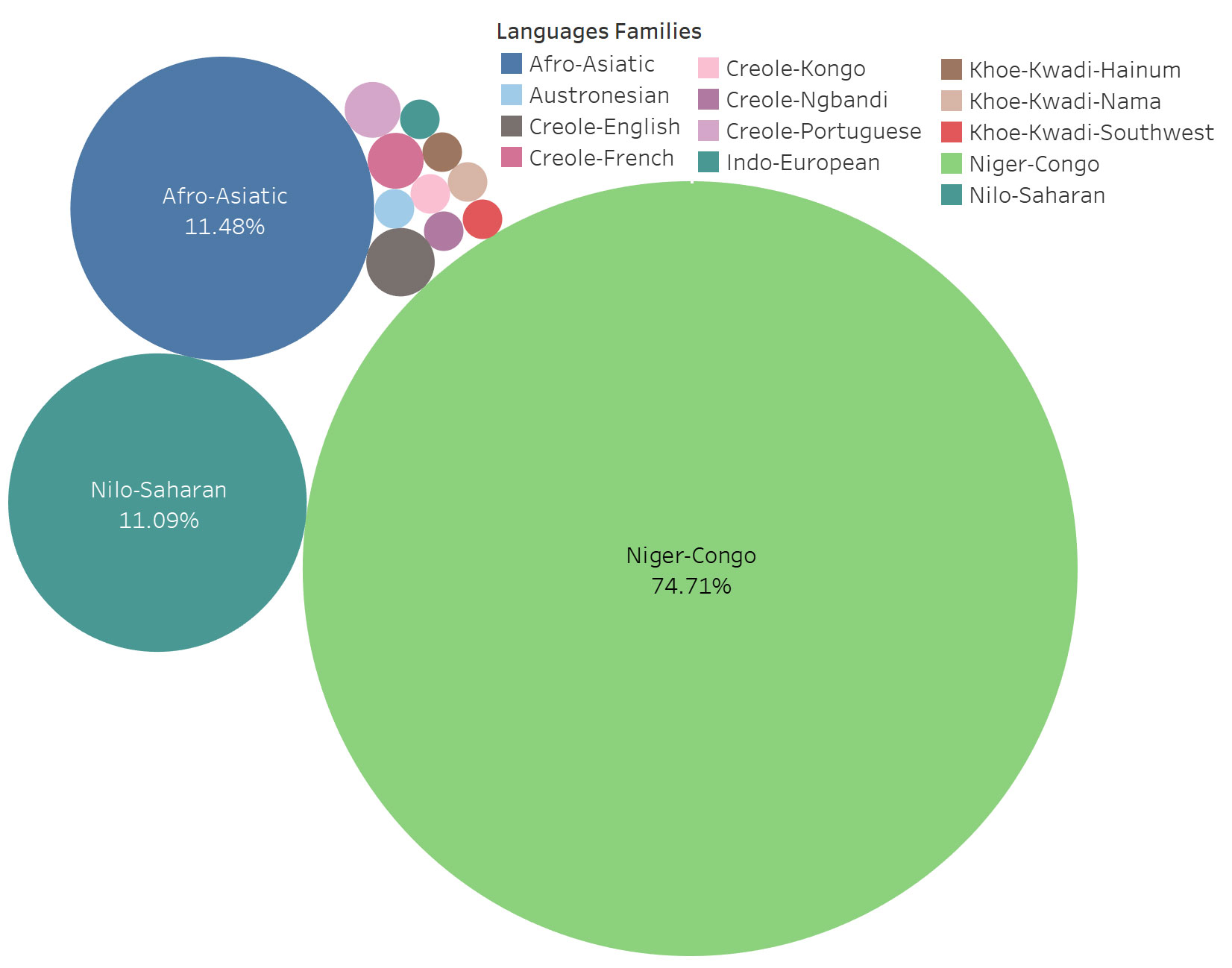}
\caption{\small Percentage of languages per family on training dataset. }
\label{fig:apx_lang_families_B} 
\end{figure*}

\ourLID~ supports $517$ African languages and language varieties. We show a large map indicating the countries and languages represented in Figure \ref{fig:map_lang_families}. Figure \ref{fig:apx_lang_families_A} and \ref{fig:apx_lang_families_B} show the number of languages covered in each country and the language family information for the languages.  We also show the languages and language codes in Table \ref{tab:lang_listI}, \ref{tab:lang_listII}, and \ref{tab:lang_listIII}.



\begin{table*}[h!]
\small
\centering
\resizebox{\textwidth}{!}{%
\begin{tabular}{ll ll ll ll }
\toprule
\textbf{ISO-3} & \textbf{Language}                & \textbf{ISO-3} & \textbf{Language}             & \textbf{ISO-3} & \textbf{Language}                       & \textbf{ISO-3} & \textbf{Language}         \\
\midrule
aar  & Afar / Qafar            & bky   & Bokyi                & dow   & Doyayo                         & gol   & Gola             \\
aba   & Abe / Abbey             & bmo   & Bambalang            & dsh   & Daasanach                      & gqr   & Gor              \\
abn   & Abua                    & bmv   & Bum                  & dua   & Douala                         & gso   & Gbaya, Southwest \\
acd   & Gikyode                 & bom   & Berom                & dug   & Chiduruma                      & gud   & Dida, Yocoboue   \\
ach   & Acholi                  & bov   & Tuwuli               & dwr   & Dawro                          & gur   & Farefare         \\
ada   & Dangme                  & box   & Bwamu / Buamu        & dyi   & Sénoufo, Djimini               & guw   & Gun              \\
adh   & Jopadhola / Adhola      & bqc   & Boko                 & dyu   & Jula                           & gux   & Gourmanchema     \\
adj   & Adjukru  / Adioukrou    & bqj   & Bandial              & ebr   & Ebrie                          & guz   & Ekegusii         \\
afr   & Afrikaans               & bsc   & Oniyan               & ebu   & Kiembu / Embu                  & gvl   & Gulay            \\
agq   & Aghem                   & bsp   & Baga Sitemu          & efi   & Efik                           & gwr   & Gwere            \\
aha   & Ahanta                  & bss   & Akoose               & ego   & Eggon                          & gya   & Gbaya, Northwest \\
ajg   & Aja                     & bst   & Basketo              & eka   & Ekajuk                         & hag   & Hanga            \\
akp   & Siwu                    & bud   & Ntcham               & eko   & Koti                           & har   & Harari           \\
alz   & Alur                    & bum   & Bulu                 & eto   & Eton                           & hau   & Hausa            \\
amh   & Amharic                 & bun   & Sherbro              & etu   & Ejagham                        & hay   & Haya             \\
ann   & Obolo                   & bus   & Bokobaru             & etx   & Iten / Eten                    & hbb   & Nya huba         \\
anu   & Anyuak / Anuak          & buy   & Bullom So            & ewe   & Ewe                            & heh   & Hehe             \\
anv   & Denya                   & bwr   & Bura Pabir           & ewo   & Ewondo                         & her   & Herero           \\
asa   & Asu                     & bwu   & Buli                 & fak   & Fang                           & hgm   & Haillom          \\
asg   & Cishingini              & bxk   & Bukusu               & fat   & Fante                          & hna   & Mina             \\
atg   & Ivbie North-Okpela-Arhe & byf   & Bete                 & ffm   & Fulfulde, Maasina              & ibb   & Ibibio           \\
ati   & Attie                   & byv   & Medumba              & fia   & Nobiin                         & ibo   & Igbo             \\
avn   & Avatime                 & bza   & Bandi                & fip   & Fipa                           & idu   & Idoma            \\
avu   & Avokaya                 & bzw   & Basa                 & flr   & Fuliiru                        & igb   & Ebira            \\
azo   & Awing                   & cce   & Chopi                & fon   & Fon                            & ige   & Igede            \\
bam   & Bambara                 & chw   & Chuabo               & fub   & Fulfulde, Adamawa              & igl   & Igala            \\
bav   & Vengo                   & cjk   & Chokwe               & fue   & Fulfulde, Borgu                & ijn   & Kalabari         \\
bba   & Baatonum                & cko   & Anufo                & fuf   & Pular                          & ikk   & Ika              \\
bbj   & Ghomala                & cme   & Cerma                & fuh   & Fulfulde, Western Niger        & ikw   & Ikwere           \\
bbk   & Babanki                 & cop   & Coptic               & ful   & Fulah                          & iqw   & Ikwo             \\
bci   & Baoule                  & cou   & Wamey                & fuq   & Fulfulde Central Eastern Niger & iri   & Rigwe            \\
bcn   & Bali                    & crs   & Seychelles Creole    & fuv   & Fulfude Nigeria                & ish   & Esan             \\
bcw   & Bana                    & csk   & Jola Kasa            & gaa   & Ga                             & iso   & Isoko            \\
bcy   & Bacama                  & cwe   & Kwere                & gax   & Oromo, Borana-Arsi-Guji        & iyx   & yaka             \\
bdh   & Baka                    & daa   & Dangaleat            & gaz   & Oromo, West Central            & izr   & Izere            \\
bds   & Burunge                 & dag   & Dagbani              & gbo   & Grebo, Northern                & izz   & Izii             \\
bem   & Bemba / Chibemba        & dav   & Dawida / Taita       & gbr   & Gbagyi                         & jgo   & Ngomba           \\
beq   & Beembe                  & dga   & Dagaare              & gde   & Gude                           & jib   & Jibu             \\
ber   & Berber                  & dgd   & Dagaari Dioula       & gid   & Gidar                          & jit   & Jita             \\
bex   & Jur Modo                & dgi   & Dagara, Northern     & giz   & South Giziga                   & jmc   & Machame          \\
bez   & Bena                    & dhm   & Dhimba               & gjn   & Gonja                          & kab   & Kabyle           \\
bfa   & Bari                    & dib   & Dinka, South Central & gkn   & Gokana                         & kam   & Kikamba          \\
bfd   & Bafut                   & did   & Didinga              & gkp   & Kpelle, Guinea                 & kbn   & Kare             \\
bfo   & Birifor, Malba          & dig   & Chidigo              & gmv   & Gamo                           & kbo   & Keliko           \\
bib   & Bisa                    & dik   & Dinka, Southwestern  & gna   & Kaansa                         & kbp   & Kabiye           \\
bim   & Bimoba                  & dip   & Dinka, Northeastern  & gnd   & Zulgo-gemzek                   & kby   & Kanuri, Manga    \\
bin   & Edo                     & diu   & Gciriku              & gng   & Ngangam                        & kcg   & Tyap             \\
biv   & Birifor, Southern       & dks   & Dinka, Southeastern  & gof   & Goofa                          & kck   & Kalanga          \\
bjv   & Bedjond                 & dnj   & Dan                  & gog   & Gogo                           & kdc   & Kutu   \\
\bottomrule
\end{tabular}%
}

\caption{\ourLID~covered Languages - Part I. }
\label{tab:lang_listI}
\end{table*}

\begin{table*}[]
\small
\centering
\resizebox{\textwidth}{!}{%
\begin{tabular}{ll ll ll ll }
\toprule
\textbf{ISO-3} & \textbf{Language}                & \textbf{ISO-3} & \textbf{Language}             & \textbf{ISO-3} & \textbf{Language}                       & \textbf{ISO-3} & \textbf{Language}         \\
\midrule
kde & Makonde               & laj & Lango                       & mfh & Matal                           & ngb & Ngbandi, Northern                     \\
kdh & Tem                   & lam & Lamba                       & mfi & Wandala                         & ngc & Ngombe                                \\
kdi & Kumam                 & lap & Laka                        & mfk & Mofu, North                     & ngl & Lomwe                                 \\
kdj & Ng’akarimojong        & lee & Lyélé                       & mfq & Moba                            & ngn & Bassa                                 \\
kdl & Tsikimba              & lef & Lelemi                      & mfz & Mabaan                          & ngo & Ngoni                                 \\
kdn & Kunda                 & lem & Nomaande                    & mgc & Morokodo                        & ngp & Ngulu                                 \\
kea & Kabuverdianu          & lgg & Lugbara                     & mgh & Makhuwa-Meetto                  & nhr & Naro                                  \\
ken & Kenyang               & lgm & Lega-mwenga                 & mgo & Meta'                           & nhu & Noone                                 \\
khy & Kele / Lokele         & lia & Limba, West-Central         & mgq & Malila                          & nih & Nyiha                                 \\
kia & Kim                   & lik & Lika                        & mgr & Mambwe-Lungu                    & nim & Nilamba / kinilyamba                  \\
kik & Gikuyu / Kikuyu       & lin & Lingala                     & mgw & Matumbi                         & nin & Ninzo                                 \\
kin & Kinyarwanda           & lip & Sekpele                     & mif & Mofu-Gudur                      & niy & Ngiti                                 \\
kiz & Kisi                  & lmd & Lumun                       & mkl & Mokole                          & nka & Nkoya / ShiNkoya                      \\
kki & Kagulu                & lmp & Limbum                      & mlg & Malagasy                        & nko & Nkonya                                \\
kkj & Kako                  & lnl & Banda, South Central        & mlr & Vame                            & nla & Ngombale                              \\
kln & Kalenjin              & log & Logo                        & mmy & Migaama                         & nnb & Nande / Ndandi                        \\
klu & Klao                  & lom & Loma                        & mnf & Mundani                         & nnh & Ngiemboon                             \\
kma & Konni                 & loq & Lobala                      & mnk & Mandinka                        & nnq & Ngindo                                \\
kmb & Kimbundu              & lot & Latuka                      & moa & Mwan                            & nse & Chinsenga                             \\
kmy & Koma                  & loz & Silozi                      & mos & Moore                           & nnw & {\color[HTML]{717171} Nuni, Southern} \\
knf & Mankanya              & lro & Laro                        & moy & Shekkacho                       & nso & Sepedi                                \\
kng & Kongo                 & lsm & Saamya-Gwe / Saamia         & moz & Mukulu                          & ntr & Delo                                  \\
knk & Kuranko               & lth & Thur / Acholi-Labwor        & mpe & Majang                          & nuj & Nyole                                 \\
kno & Kono                  & lto & Tsotso                      & mpg & Marba                           & nus & Nuer                                  \\
koo & Konzo                 & lua & Tshiluba                    & mqb & Mbuko                           & nwb & Nyabwa                                \\
koq & Kota                  & luc & Aringa                      & msc & Maninka, Sankaran               & nxd & Ngando                                \\
kqn & Kikaonde              & lue & Luvale                      & mur & Murle                           & nya & Chichewa                              \\
kqp & Kimré                 & lug & Luganda                     & muy & Muyang                          & nyb & Nyangbo                               \\
kqs & Kisi                  & lun & Lunda                       & mwe & Mwera                           & nyd & Olunyole / Nyore                      \\
kqy & Koorete               & luo & Dholuo / Luo                & mwm & Sar                             & nyf & Giryama                               \\
kri & Krio                  & lwg & Wanga                       & mwn & Cinamwanga                      & nyk & Nyaneka                               \\
krs & Gbaya                 & lwo & Luwo                        & mws & Mwimbi-Muthambi                 & nym & Nyamwezi                              \\
krw & Krahn, Western        & maf & Mafa                        & myb & Mbay                            & nyn & Nyankore / Nyankole                   \\
krx & Karon                 & mas & Maasai                      & myk & Sénoufo, Mamara                 & nyo & Nyoro                                 \\
ksb & Shambala / Kishambala & maw & Mampruli                    & myx & Masaaba                         & nyu & Nyungwe                               \\
ksf & Bafia                 & mbu & Mbula-Bwazza                & mzm & Mumuye                          & nyy & Nyakyusa-Ngonde / Kyangonde           \\
ksp & Kabba                 & mck & Mbunda                      & mzw & Deg                             & nza & Mbembe, Tigon                         \\
ktj & Krumen, Plapo         & mcn & Masana / Massana            & naq & {\color[HTML]{717171} Khoekhoe} & nzi & Nzema                                 \\
ktu & Kikongo               & mcp & Makaa                       & naw & Nawuri                          & odu & Odual                                 \\
kua & Oshiwambo             & mcu & Mambila, Cameroon           & nba & Nyemba                          & ogo & Khana                                 \\
kub & Kutep                 & mda & Mada                        & nbl & IsiNdebele                      & oke & Okpe                                  \\
kuj & Kuria                 & mdm & Mayogo                      & ncu & Chunburung                      & okr & Kirike                                \\
kus & Kusaal                & mdy & Maale                       & ndc & Ndau                            & oku & Oku                                   \\
kvj & Psikye                & men & Mende                       & nde & IsiNdebele                      & orm & Oromo                                 \\
kwn & Kwangali              & meq & Merey                       & ndh & Ndali                           & ozm & Koonzime                              \\
kyf & Kouya                 & mer & Kimiiru                     & ndj & Ndamba                          & pcm & Nigerian Pidgin                       \\
kyq & Kenga                 & mev & Maan / Mann                 & ndo & Ndonga                          & pem & Kipende                               \\
kzr & Karang                & mfe & Morisyen / Mauritian Creole & ndv & Ndut                            & pkb & Kipfokomo / Pokomo                    \\
lai & Lambya                & mfg & Mogofin                     & ndz & Ndogo    \\
\bottomrule
\end{tabular}%
}
\caption{\ourLID~covered Languages - Part II }
\label{tab:lang_listII}
\end{table*}

\begin{table*}[]
\small
\centering
\resizebox{0.8\textwidth}{!}{%
\begin{tabular}{ll ll ll}
\toprule
\textbf{ISO-3} & \textbf{Language}                & \textbf{ISO-3} & \textbf{Language}             & \textbf{ISO-3} & \textbf{Language}                                \\
\midrule
pov & Guinea-Bissau Creole      & tcd & Tafi               & won                      & Wongo               \\
poy & Pogolo / Shipogoro-Pogolo & ted & Krumen, Tepo       & xan                      & Xamtanga            \\
rag & Lulogooli                 & tem & Timne              & xed                      & Hdi                 \\
rel & Rendille                  & teo & Teso               & xho                      & Isixhosa            \\
rif & Tarifit                   & tex & Tennet             & xnz                      & Mattokki            \\
rim & Nyaturu                   & tgw & Senoufo, Tagwana   & xog                      & Soga                \\
rnd & Uruund                    & thk & Tharaka            & xon                      & Konkomba            \\
rng & Ronga / ShiRonga          & thv & Tamahaq, Tahaggart & xpe                      & Kpelle              \\
rub & Gungu                     & tir & Tigrinya           & xrb                      & Karaboro, Eastern   \\
run & Rundi / Kirundi           & tiv & Tiv                & xsm                      & Kasem               \\
rwk & Rwa                       & tke & Takwane            & xtc                      & Katcha-Kadugli-Miri \\
sag & Sango                     & tlj & Talinga-Bwisi      & xuo                      & Kuo                 \\
saq & Samburu                   & tll & Otetela            & yal                      & Yalunka             \\
sba & Ngambay                   & tog & Tonga              & yam                      & Yamba               \\
sbd & Samo, Southern            & toh & Gitonga            & yao                      & Yao / Chiyao        \\
sbp & Sangu                     & toi & Chitonga           & yat                      & Yambeta             \\
sbs & Kuhane                    & tpm & Tampulma           & yba                      & Yala                \\
sby & Soli                      & tsc & Tshwa              & ybb                      & Yemba               \\
sef & Sénoufo, Cebaara          & tsn & Setswana           & yom                      & Ibinda              \\
ses & Songhay, Koyraboro Senni  & tso & Tsonga             & yor                      & Yoruba              \\
sev & Sénoufo, Nyarafolo        & tsw & Tsishingini        & yre                      & Yaoure              \\
sfw & Sehwi                     & ttj & Toro / Rutoro      & zaj                      & Zaramo              \\
sgw & Sebat Bet Gurage          & ttq & Tawallammat        & zdj                      & Comorian, Ngazidja  \\
shi & Tachelhit                 & ttr & Nyimatli           & zga                      & Kinga               \\
shj & Shatt                     & tui & Toupouri           & ziw                      & Zigula              \\
shk & Shilluk                   & tul & Kutule             & zne                      & Zande / paZande     \\
sid & Sidama                    & tum & Chitumbuka         & zul                      & Isizulu             \\
sig & Paasaal                   & tuv & Turkana            &  &                     \\
sil & Sisaala, Tumulung         & tvu & Tunen              &  &                     \\
sna & Shona                     & twi & Twi                &  &                     \\
snf & Noon                      & umb & Umbundu            &  &                     \\
sng & Sanga / Kiluba            & urh & Urhobo             &  &                     \\
snw & Selee                     & uth & ut-Hun            &  &                     \\
som & Somali                    & vag & Vagla              &  &                     \\
sop & Kisonge                   & vai & Vai                &  &                     \\
sor & Somrai                    & ven & Tshivenda          &  &                     \\
sot & Sesotho                   & vid & Chividunda         &  &                     \\
soy & Miyobe                    & vif & Vili               &  &                     \\
spp & Senoufo, Supyire          & vmk & Makhuwa-Shirima    &  &                     \\
ssw & Siswati                   & vmw & Macua              &  &                     \\
suk & Sukuma                    & vun & Kivunjo            &  &                     \\
sus & Sosoxui                   & vut & Vute               &  &                     \\
swa & Swahili                   & wal & Wolaytta           &  &                     \\
swc & Swahili Congo             & wbi & Vwanji             &  &                     \\
swh & Swahili                   & wec & Guere              &  &                     \\
swk & Sena, Malawi              & wes & Pidgin, Cameroon   &  &                     \\
sxb & Suba                      & wib & Toussian, Southern &  &                     \\
taq & Tamasheq                  & wmw & Mwani              &  &                     \\
tcc & Datooga                   & wol & Wolof              &  & 
\\
\bottomrule
\end{tabular}%
}

\caption{\ourLID~covered Languages - Part III.}
\label{tab:lang_listIII}
\end{table*}

\label{sec:appendix}
\end{document}